\let\oldyear\year
\pdfoutput=1
\documentclass[figuresright]{ieeeaccess}

\let\year\oldyear

\usepackage{cite}
\usepackage{amsmath,amssymb,amsfonts}
\usepackage{algorithmic}
\usepackage{graphicx}
\usepackage{float}
\usepackage{textcomp}
\usepackage{url}
\usepackage{pdflscape}
\usepackage{svg}
\usepackage{rotating}   
\usepackage{adjustbox}
\usepackage{foreststyle}
\definecolor{revised}{RGB}{0, 0, 0}
\usepackage{eurosym}

\usepackage{tikz}
\NewSpotColorSpace{PANTONE}
\AddSpotColor{PANTONE} {PANTONE3015C} {PANTONE\SpotSpace 3015\SpotSpace C} {1 0.3 0 0.2}
\SetPageColorSpace{PANTONE}
\definecolor{accessblue}{cmyk}{1, 0.3, 0, 0.2}
\definecolor{greycolor}{cmyk}{0,0,0,.8}

\usepackage{booktabs}
\usepackage{adjustbox}
\usepackage[edges]{forest}
\usepackage{gensymb}
\usetikzlibrary{arrows.meta,positioning,shapes.geometric}

\tikzset{
    decision/.style={ellipse, draw=gray, fill=#1, inner sep=2pt, minimum width=2cm, font=\small\bfseries\sffamily},
    action/.style={rectangle, rounded corners=3pt, draw=gray, fill=#1, inner sep=4pt, minimum width=1cm, font=\small\bfseries\sffamily},
    charge/.style={fill=green!80!white},
    discharge/.style={fill=green!80!black, text=white},
    legend box/.style={rectangle, draw=gray!50, minimum size=8pt, inner sep=0pt}
}

\usepackage{bm}
\makeatletter
\AtBeginDocument{\DeclareMathVersion{bold}
\SetSymbolFont{operators}{bold}{T1}{times}{b}{n}
\SetSymbolFont{NewLetters}{bold}{T1}{times}{b}{it}
\SetMathAlphabet{\mathrm}{bold}{T1}{times}{b}{n}
\SetMathAlphabet{\mathit}{bold}{T1}{times}{b}{it}
\SetMathAlphabet{\mathbf}{bold}{T1}{times}{b}{n}
\SetMathAlphabet{\mathtt}{bold}{OT1}{pcr}{b}{n}
\SetSymbolFont{symbols}{bold}{OMS}{cmsy}{b}{n}
\renewcommand\boldmath{\@nomath\boldmath\mathversion{bold}}}
\makeatother

\DeclareFontFamilySubstitution{TS1}{times}{cmr}
\DeclareFontShape{TS1}{times}{n}{n}{<->ssub * cmr/m/n}{}

\DeclareFontFamily{T1}{pcr}{}
\DeclareFontShape{T1}{pcr}{m}{n}{<->ssub * cmtt/m/n}{}

\def\BibTeX{{\rm B\kern-.05em{\sc i\kern-.025em b}\kern-.08em
    T\kern-.1667em\lower.7ex\hbox{E}\kern-.125emX}}

\begin{document}
\history{Date of publication xxxx 00, 0000, date of current version xxxx 00, 0000.}
\doi{xxxx}

\title{Explainable Data-driven Deep Reinforcement Learning Methods for Optimal Energy Management in Buildings}
\author{\uppercase{Hallah Shahid Butt}\authorrefmark{1}, 
\uppercase{Qiong Huang}\authorrefmark{1}, \uppercase{G{\"o}khan Demirel}\authorrefmark{1}, \uppercase{Kevin F{\"o}rderer}\authorrefmark{1}, \uppercase{Erfan Tajalli-Ardekani}\authorrefmark{1}, \uppercase{Simon Waczowicz}\authorrefmark{1}, \uppercase{Luigi Spatafora}\authorrefmark{1}, \uppercase{Veit Hagenmeyer}\authorrefmark{1}, and \uppercase{Benjamin Sch{\"a}fer}\authorrefmark{1}}
% \IEEEmembership{Member, IEEE}}

% For submission system: orcid
% Hallah Shahid Butt: \orcid{0009-0001-8045-4883}
% Qiong Huang: \orcid{0000-0002-1958-6094}
% G{\"o}khan Demirel: \orcid{0000-0002-1234-5474}
% Kevin F{\"o}rderer: \orcid{0000-0002-9115-670X}
% ...
% Veit Hagenmeyer: \orcid{0000-0002-3572-9083}
% Benjamin Sch{\"a}fer: \orcid{0000-0003-1607-9748}

\address[1]{Karlsruhe Institute of Technology, 76344 Eggenstein-Leopoldshafen, Germany}

% \tfootnote{This paragraph of the first footnote will contain support
% information, including sponsor and financial support acknowledgment. For
% example, ``This work was supported in part by the U.S. Department of
% Commerce under Grant BS123456.''}

\markboth
% {Author \headeretal: Preparation of Papers for IEEE TRANSACTIONS and JOURNALS}
% {Author \headeretal: Preparation of Papers for IEEE TRANSACTIONS and JOURNALS}
{Preprint submitted to IEEE Access}
{Preprint submitted to IEEE Access}

\corresp{Corresponding author: Hallah Shahid Butt (e-mail: hallah.butt@kit.edu).}

\begin{abstract}
The increasing integration of renewable energy sources into power systems, particularly in buildings equipped with photovoltaic (PV) panels and energy storage systems, introduces significant complexity in energy systems. Volatile power generation, varying electricity tariffs, and increased entities, e.g., PV systems, and heat pumps,  have increased the complexity and made the system harder to operate. This leads to the demand for additional control and optimization routes including data-based controls, such as reinforcement learning. While deep reinforcement learning (DRL) has emerged as a promising solution to optimize building operations in dynamic and ever more complex environments, its black-box nature impedes user trust and practical adoption. This paper presents a framework for explainable deep reinforcement learning (XRL) applied to energy management in residential buildings. We demonstrate its usage on both synthetic data but also on real-world data from the Living Lab Energy Campus (LLEC) at KIT. 
We train and compare both on-policy and off-policy DRL agents on an expanded state space that incorporates real-time measurements (demand, PV generation, battery power, state of charge), external signals (dynamic electricity price, local weather data), calendrical and holiday indicators, and forecasts for demand and price. 
Our experimental results indicate that on-policy algorithms, particularly Advantage Actor Critic (A2C) and Proximal Policy Optimization (PPO), outperform off-policy methods in terms of cumulative rewards and policy stability. 
To explain these models, we employ post-hoc interpretation techniques to elaborate the learned control policies. 
Our findings demonstrate that the XRL framework not only reduces electricity costs through optimal battery management, but also provides transparent, actionable insights into the agent's decision-making process. 
\end{abstract}

\begin{keywords}
Battery Storage Optimization, Building Energy Management, Deep Reinforcement Learning, Explainable AI, Renewable Energy Integration
\end{keywords}

\titlepgskip=-21pt

\maketitle

\section{Introduction}
\label{sec:introduction}
\PARstart{G}{lobal} energy consumption has grown rapidly in recent decades due to urbanization and technological advancement \cite{thirunavukkarasu2022role}. Renewable and distributed energy resources (DERs) are increasingly integrated into power systems 
% as sustainable alternatives 
\cite{butt2024explainable}. 
This large-scale adoption of DERs introduces new challenges: their geographical dispersion and weather dependence increase grid complexity and uncertainty \cite{shengren2023optimal}. 
For example, solar generation in Germany varies seasonally, with an average of 262 sunny hours in August compared to only 42 in December \cite{hoursGermany}. 
Such fluctuations highlight the need for advanced optimization and control strategies to balance supply and demand. At the building scale, the integration of local generation, on-site storage, and participation in real-time electricity markets further complicates energy management. Moreover, the growing usage of power-based devices, such as electric vehicles and heat pumps, strengthens the coupling between thermal and electrical sectors, increasing the complexity of optimization and control at both the building and grid levels.

%%introducing MPC here 
These increasing levels of complexity in building energy systems necessitate advanced control approaches capable of handling multi-domain interactions, uncertainty, and operational constraints. Model Predictive Control (MPC) has emerged as a promising strategy for such applications, offering the ability to optimize energy flows over a prediction horizon while accounting for system dynamics and forecast information \cite{Afram2014, Killian2016, Ma2012}. In building power management, MPC enables coordinated operation of local generation, storage, and flexible loads to minimize energy costs, reduce peak demand, and improve overall efficiency \cite{Drgona2020, Sturzenegger2016}. Furthermore, its capability to incorporate forecasts of renewable generation, occupancy, and market signals makes MPC particularly suitable for real-time and demand-responsive energy management systems \cite{Oldewurtel2013, Zhang2018}.

While MPC relies on explicit system models and forecasts to optimize future control actions, its performance strongly depends on model accuracy and computational tractability, particularly in large-scale and uncertain environments \cite{Drgona2020, Killian2016}. In contrast, Reinforcement Learning (RL) provides an AI-based control approach that learns policies directly from interaction with the environment, without requiring an explicit system model or precise forecasts \cite{Wei2022, Zhang2021}. By continuously adapting to changing operating conditions and uncertainties, RL can outperform MPC in scenarios where accurate models are difficult to obtain or where system dynamics evolve over time \cite{Chen2021}. Moreover, recent studies demonstrate that RL-based controllers can effectively handle nonlinearities, stochastic disturbances, and multi-objective optimization, making them well-suited for real-time energy management in buildings with distributed energy resources and flexible loads \cite{Gao2020, Yu2023, demirel2025morl}. Consequently, RL represents a promising direction for developing scalable and autonomous control frameworks that complements or even surpasses traditional MPC approaches in complex energy systems.

A wide range of approaches has been proposed for energy management, from traditional controllers such as thermostats and PID regulators to embedded systems and automated devices \cite{OUEDRAOGO2023126922}. 
More recently, artificial intelligence (AI) and machine learning (ML) methods, including support vector machines, random forests, and neural networks, have demonstrated significant potential in improving building energy efficiency \cite{hossain2023review}. 
RL has emerged as particularly promising, as it continuously interacts with the environment to derive optimal energy management policies, enabling cost-effective and adaptive control \cite{rojek2025internet}. 
Despite this potential, ML and RL methods are often criticized for their “black-box” nature. A deeper understanding of model limitations can help AI developers enhance transparency and robustness, while increasing domain experts’ trust in model outputs and willingness to act on their recommendations.
Explainable AI (XAI), particularly post-hoc explanation methods, is therefore essential for increasing transparency and facilitating adoption in energy management systems \cite{beechey2023explaining}.

% The present paper addresses these challenges by developing reinforcement learning–based policies for building energy optimization and enhancing their interpretability through post-hoc explanation techniques. 
% The main contributions are:
% \begin{enumerate}
% \item Introduction of RL-based policies to optimize building energy consumption.
% \item Integration of post-hoc explanations to improve transparency and user trust in learned policies.
% % \item An explainability framework that provides transparent insights into the trained RL policies, thereby improving user understanding and trust.
% \end{enumerate}

This paper addresses these explainability challenges by developing a unified framework that bridges the gap between black-box DRL control and domain-specific energy requirements. Unlike existing studies that treat explainability as an auxiliary step, this work focuses on the functional integration of post-hoc tools within the operational logic of building energy systems. Our main contributions are as follows:
\begin{enumerate}
    \item We propose a structured and generalized integrated pipeline that links DRL policy outputs to complementary explainability mechanisms, including decision-tree–based policy approximation and feature ablation analysis, specifically adapted for high-dimensional building energy state spaces.
    \item We establish a comparative evaluation protocol across multiple Deep Reinforcement Learning (DRL) algorithms (e.g., PPO, A2C), enabling consistent benchmarking under identical environmental and operational constraints.
    \item We introduce a state-space design of the RL models that incorporates both historical measurements and short-term forecast information (e.g., demand and electricity price signals), with forecast horizon selection validated through quantitative error analysis to ensure operational relevance.
    % \item We implement a unified DRL-based building energy management framework capable of coordinating distributed energy resources, storage systems, and flexible loads under real-time market conditions.
    \item Through the integrated explainability analysis, we quantitatively identify the dominant state variables influencing policy formation and analyze how forecast information and system uncertainty shape learned control strategies.
\end{enumerate}

The first two points represent the methodological contributions of our work, while the third point represents the system-level contribution of the work. Lastly, one is related to insight contributions.

The remainder of this paper is structured as follows: 
Section \ref{section-lr} reviews related works. 
Section \ref{section-methodology} presents the methodology. 
Section \ref{section-results} discusses the results and interpretability analysis, and Section \ref{section-conclude} concludes the study with future directions.
\begin{figure*}[htbp]
    \centering
    \includegraphics[width=1.0\linewidth, keepaspectratio]{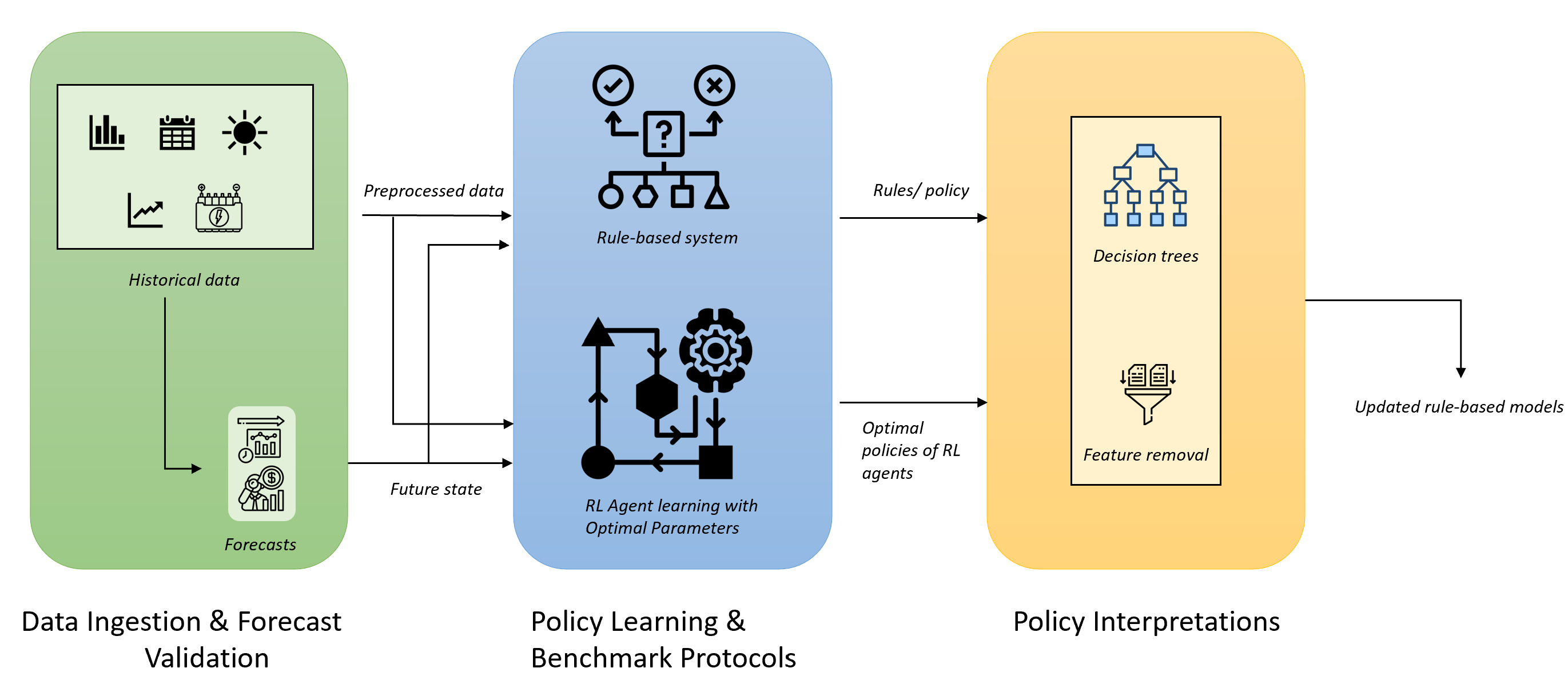}
    \caption{Overall methodological framework using reinforcement learning for energy system optimization. Historical and forecast data are preprocessed to generate the future state inputs for RL and rule-based models. The reinforcement learning agents produce operational policies, which are iteratively refined using explainable AI (XAI) techniques such as decision tree analysis and feature removal to enhance interpretability and model performance.}
    \label{fig:methodology}
\end{figure*}

\section{Literature Review}
\label{section-lr}
Machine learning (ML), particularly deep learning, has significantly advanced energy system optimization by improving forecasting accuracy, and system reliability \cite{aiml, su14084832}. 
Achieving net-zero emission targets requires efficient demand–supply balancing, yet non-existent or poorly performing building energy management remains a significant challenge, particularly in older, energy-inefficient buildings \cite{IEA2021NetZero,IEA2023Buildings}. Consequently, building energy management systems (BEMS) have become a focal point of recent research due to their potential to improve operational efficiency, reduce emissions, and support demand flexibility \cite{Drgona2020,Wei2022,Xu2023BEMSReview}.
Survey studies highlight that AI and ML approaches, especially supervised learning methods, are widely applied to optimize energy use in buildings \cite{en17174277}.

%%contrast with other ML subcategories instead of current 1st sentence ->to control and optimize building supervised learning is not very helpful --
With their ability to capture arbitrary and potentially nonlinear relationships in large datasets, deep learning methods further enhance these capabilities. 
Reinforcement learning (RL), and more recently deep reinforcement learning (DRL), has been applied to building energy management, enabling adaptive and data-driven decision-making in real-world environments \cite{Drgona2020, Wei2022, Zhang2021}.
Applications of DRL span power system operations, renewable energy optimization, and building-level demand response, showing strong potential for efficiency gains and cost reductions \cite{shengren2023optimal, cao2020reinforcement, demirel2025hp, PINCIROLI2022752, zhang2019deep}.

Traditional MPC methods remain the most commonly used approaches, particularly in balancing occupant comfort with energy efficiency. 
Recent work has, for example, applied economic MPC to occupant-oriented demand response in buildings, jointly considering grid-supportive behavior and thermal comfort \cite{Frahm2022EMPC}. 
In addition to such model-based approaches, price-based, rule-based control schemes have been proposed for demand response and peak shaving in building clusters, such as the Extended Price Storage Control+ (EPSC+) for coordinating modulating heat pumps under dynamic pricing \cite{Langner2024EPSC}. 
Furthermore, a systematic comparison of deep RL algorithms against rule-based controllers and MPC for residential heat pump operation under dynamic tariffs has demonstrated that the best RL agent can outperform rule-based methods and approach MPC-level performance without requiring an explicit thermal model~\cite{demirel2025hp}.
In parallel, recent studies have explored generalizable reinforcement learning frameworks for HVAC control using real-world data, demonstrating strong adaptability but also highlighting data availability and interpretability challenges \cite{RLGeneralizable2024}. 
However, MPC strategies are typically building-specific and must be reconfigured when structural characteristics, appliances, or occupancy patterns change \cite{liu2020optimization, WANG2023120430}. 
By contrast, DRL policies offer greater adaptability, allowing flexible and cost-efficient solutions across diverse and evolving environments. 
A practical limitation, however, is that most available BEMS datasets are synthetic. RL agents trained on these datasets often fail to generalize to real-world conditions, where empirical residential data is scarce and difficult to share due to privacy concerns \cite{wang2023rl4rs}.
This underscores the importance of approaches validated on real residential datasets. 

Existing RL approaches are mostly based on purely synthetic datasets and do not allow a straightforward practical implementation \cite{yu2021review}. Their synthetic nature omits the system dynamics and reward structures of realistic operational datasets. This gap encouraged us to study not only a synthetic test case  but also empirical data that captures the operational dynamics of real-world buildings \cite{osti_1395882}. 

Moreover, another critical limitation in current research is the lack of interpretability. 
While RL has demonstrated strong performance in energy management, its black-box nature raises concerns about trustworthiness and accountability. 
XAI has been increasingly recognized as essential for enhancing transparency, regulatory compliance, and user confidence in AI-driven systems \cite{gerlings2020reviewing, towardsxai}. 
Existing studies have explored interpretability in energy applications \cite{en14217367, Machlev2022, WENNINGER2022118300}, but few works integrate XAI directly with RL to explain decision-making processes \cite{puiutta2020explainable}. These explanations in building energy management systems aim not only to improve model transparency but also to provide actionable insights that enable building operators and stakeholders to make informed operational and energy-efficiency decisions \cite{TEIXEIRA2025116246}.
This gap motivates the present study, which investigates both on-policy and off-policy RL methods for building energy management and complements them with post-hoc explanations to improve interpretability. 

\section{Methodology}
\label{section-methodology}
The objective of this study is to develop RL–based policies for building energy management that are both cost-effective and interpretable. This section presents the structured explainable DRL framework proposed in this work, which consists of three interconnected modules: (i) data ingestion and forecast validation, (ii) policy learning and benchmark protocol, and (iii) policy interpretations. 
The overall framework of the methodology is shown in Fig.~\ref{fig:methodology}.
% \begin{figure*}
%     \centering
%     \includesvg[width=1.0\linewidth, keepaspectratio]{methodology final}
%     \caption{Workflow for forecasting electricity prices and demand for the next one hour.}
%     \label{fig:methodology-a}
% \end{figure*}
% \begin{figure*}
%     \centering
%     \includesvg[width=1.0\linewidth, keepaspectratio]{method(1)}
%     \caption{Framework to optimize the energy in a building using XRL. RL policy is generated after training, followed by the interpretations using decision trees.}
%     \label{fig:methodology}
% \end{figure*}

% \includepdf{methodology.pdf}
\begin{figure}[ht]
    \centering
    \includesvg[width=1.0\linewidth]{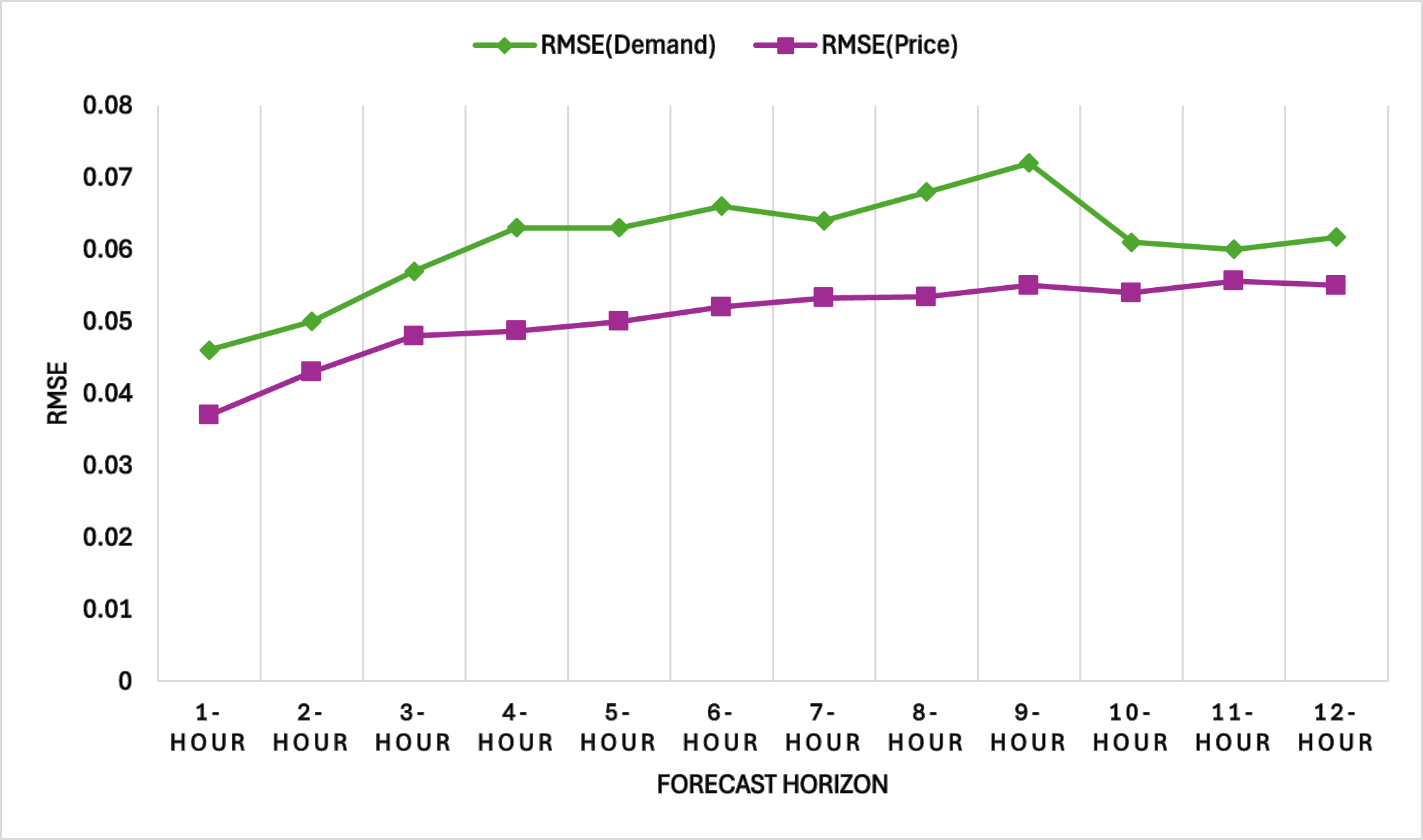}
    \caption{Comparative analysis of RMSE calculated at different levels of forecasts for demand and price variables.}
    \label{fig:rmse}
\end{figure}
\begin{figure*}[t]
    \centering

    \begin{minipage}{\textwidth}
        \centering
        \includegraphics[width=\linewidth]{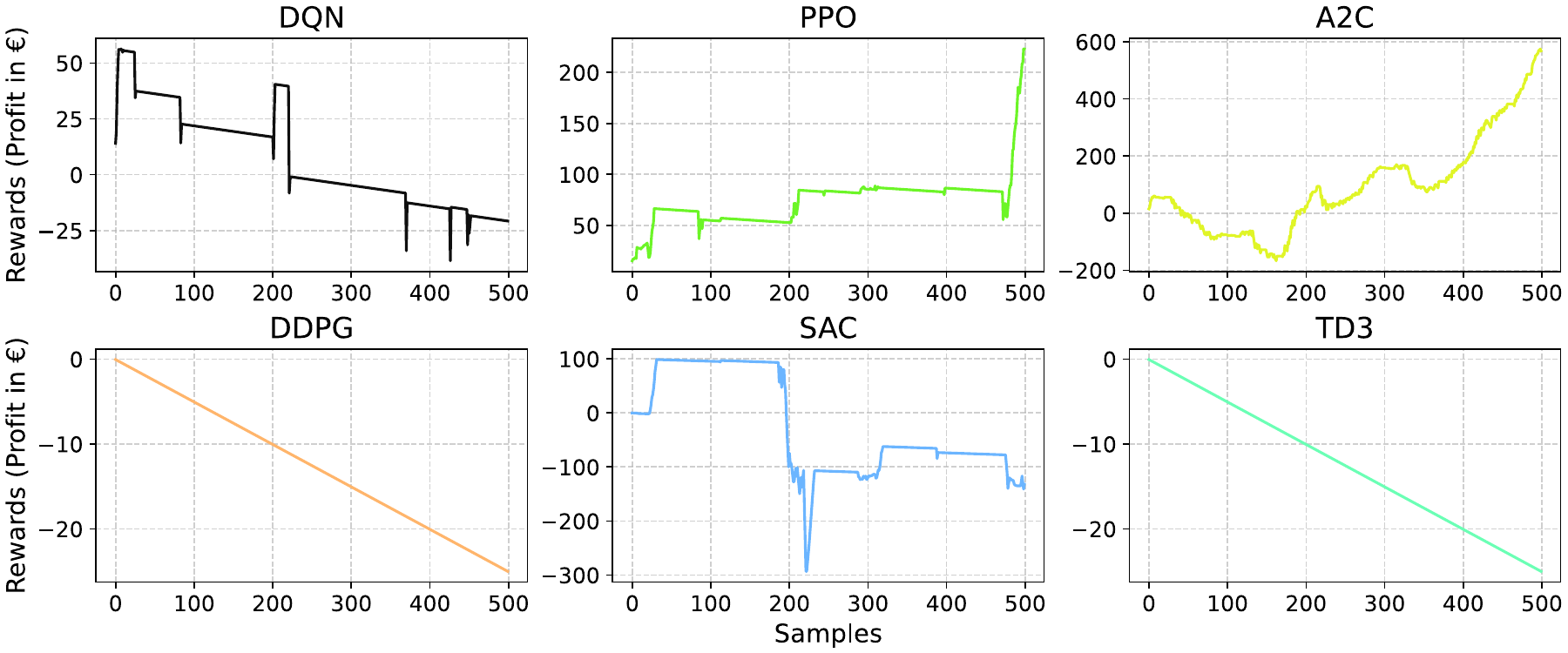}
        \label{fig:cummreward-a2c}
        \small (a) Cumulative reward plots of all RL agents on test dataset. A2C is performing best among other RL methods. These profit gains are compared with the baseline reward, which is \textbf{0}.
    \end{minipage}

    \vspace{0.5em}

    \begin{minipage}{\textwidth}
        \centering
        \includegraphics[width=\linewidth]{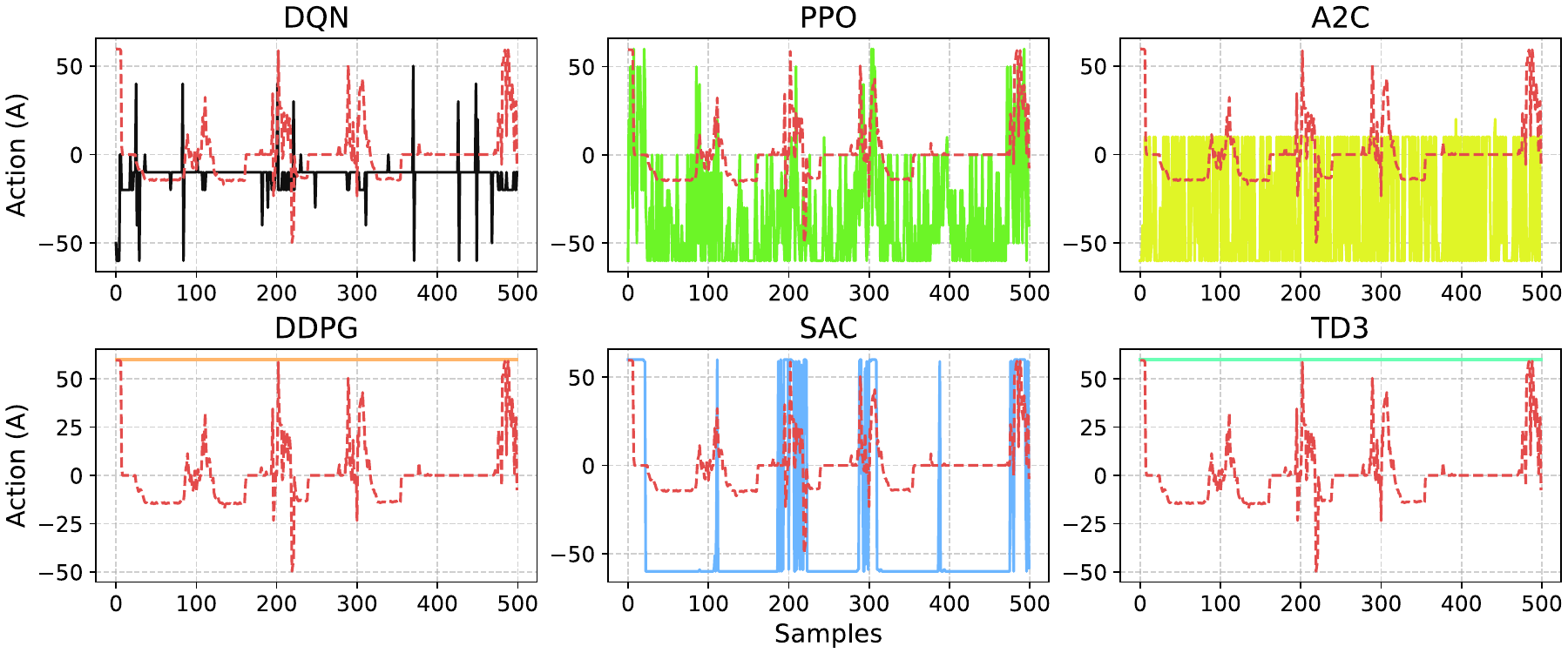}
        \label{fig:actionsnew}
        \small (b) Actions taken by RL agents and their comparison with actual battery's current (red dashed lines). This comparative analysis show that A2C takes more effective actions to keep future states in consideration and gains maximum profit.
    \end{minipage}

    \caption{Cumulative rewards and corresponding actions obtained by different reinforcement learning agents. The upper two rows present the cumulative rewards, expressed as profit gains (in euros), while the lower two rows show the battery idle, charging and discharging actions (in amperes). The A2C agent exhibits the highest cumulative reward, suggesting a more effective policy for optimizing energy-related profit.}
    \label{fig:both_images}
\end{figure*}

\subsection{Data Ingestion and Forecast Validation}
This module is responsible for preparing the dataset for training within the RL framework. It includes data preprocessing steps as well as the forecasting of key variables, such as load demand and electricity prices. The experiments are conducted using data from the Living Lab Energy Campus (LLEC) \cite{llec}. Measurements were collected between 24 June 2024 and 25 September 2024, covering periods of high solar irradiation. 
The dataset includes:
\begin{enumerate}
    \item Power-related data: residential demand, photovoltaic (PV) generation, and electricity purchased from the grid.
    \item Weather information: temperature, wind speed, air pressure, and humidity.
\end{enumerate}

Dynamic electricity prices are incorporated using the ENTSO-e platform \cite{entsoe}. 
Missing values in the dataset are imputed using a forward-fill method, a common time-series imputation technique that propagates the last observed value forward \cite{Hyndman2018Forecasting}. Future price and demand values with a prediction horizon of up to one hour are forecast using a Long Short-Term Memory (LSTM) network, which is well suited for capturing nonlinear temporal dependencies in sequential data \cite{Hochreiter1997}. Multiple forecast horizons are evaluated, and their accuracy is quantified using the root mean square error (RMSE). A comparative analysis is performed for both target variables (demand and electricity price). As shown in Fig.~\ref{fig:rmse}, the 1-hour-ahead horizon provides the best predictive accuracy and stability. 
This validation ensures that including forecast information is beneficial to the state representation without introducing additional variability into the learning process.% Therefore, the selection of the 1-hour horizon constitutes a design decision grounded in predictive reliability rather than an arbitrary parameter choice.

For training RL model, we split the dataset into 80\% training and 20\% testing. After training, the policy is evaluated on test dataset. 
% Household demand profile during the measurement period is shown in Figure \ref{fig:demand}. 

\subsection{Policy Learning and Benchmark Protocol}
We train six RL algorithms to develop and evaluate our control agents: Deep Q-Network (DQN) \cite{Mnih2015}, Proximal Policy Optimization (PPO) \cite{Schulman2017}, Advantage Actor–Critic (A2C) \cite{Mnih2016}, Soft Actor–Critic (SAC) \cite{Haarnoja2018}, Twin Delayed Deep Deterministic Policy Gradient (TD3) \cite{Fujimoto2018}, and Deep Deterministic Policy Gradient (DDPG) \cite{Lillicrap2015}. These algorithms are selected to cover a broad range of learning paradigms, including both on-policy (PPO, A2C) and off-policy (DQN, SAC, TD3, DDPG) methods. The inclusion of both on-policy and off-policy methods establishes a structured benchmarking suite, allowing assessment of policy stability, sample efficiency, and sensitivity to stochastic dynamics under identical environmental conditions. As is often the case for existing complex control systems, we do not have access to a simple representation of the actual control logic underlying the empirical data. Hence, we approximate the behavior of the existing system using a decision tree trained on the empirical dataset to capture its decision-making patterns in a rule-based form.

We adopt an automated hyperparameter optimization approach using the Optuna library \cite{optuna}. The search range of each hyperparameter for all algorithms is stated in Table.~\ref{tab:searchrange}. In parallel, 8 environments were trained and the optimal set of hyperparameters were selected for further training. Consequently, we selected the best-performing hyperparameters with best mean reward to reduce manual tuning effort.

All ML experiments were conducted on High-Performance Computing (HPC) infrastructure at KIT, specifically from the Helmholtz AI computing resources (HAICORE). These include A100 GPUs to support researchers within Helmholtz AI community \cite{haicore_kit_2026}.
%\textcolor{revised}{These experiments are conducted at the high speed Helmholtz AI computing resources (HAICORE) provided by Helmholtz AI community. These resources are used for light-weight projects to support researchers within Helmholtz AI community \cite{haicore_kit_2026}. No GPU acceleration or external cloud resources were used. }

\subsubsection{Reinforcement Learning Environment}
The considered environment of our empirical dataset is a building located on a research campus, primarily operated as an office facility, and equipped with photovoltaic (PV) panels and battery storage \cite{llec}. Consequently, its energy demand profile is characterized by higher consumption during working hours and lower demand in the evenings, on weekends and during holidays.
The total household demand is met through a combination of PV generation $P_{\text{pv}}$, battery charge $P_{\text{battery}}$, and electricity purchased from the grid $P_{\text{grid}}$. 
The power balance is defined as:
\begin{equation}
    P_{\text{grid}} = P_{\text{pv}} + P_{\text{battery}} - P_{\text{demand}} % P_{\text{demand}} ?
\end{equation}
The RL agent controls the battery charging/discharging while respecting physical constraints (e.g., avoiding overcharging or deep discharging).

\begin{itemize}
    \item \textbf{State Space}: The selection of the state space is a critical component in RL-based energy management environments, as it directly influences policy convergence and overall learning performance. In this study, the state representation integrates both historical observations and short-term predictive information to enhance the agent’s decision-making capability under dynamic operating conditions. A detailed description of the state variables is provided in Table~\ref{tab:statespace}.
    
% \textcolor{revised}{To determine the appropriate forecasting horizon for electricity demand and market price signals, multiple prediction intervals ranging from 1 to 12 hours ahead were evaluated. Forecast accuracy was assessed using the root mean square error (RMSE) metric by comparing predicted values against actual measurements. The results indicated that the 1-hour-ahead forecasts consistently achieved the lowest RMSE. Consequently, a 1-hour forecasting horizon was adopted, as it provides a favorable trade-off between prediction accuracy and decision-making relevance within the RL framework.}
\begin{table}[ht]
    \centering
    \resizebox{\columnwidth}{!}{
    \begin{tabular}{ccccccc}
    \toprule
       Hyperparameter  & DQN & PPO & A2C & SAC & DDPG & TD3 \\
       \midrule \\
       Seed  & [1,..,10] & [1,..,20] & [1,..,20] & [1,..,20] & [1,..,20] & [1,..,20] \\
       Gamma & [0.89,..,0.99] & [0.89,..,0.99] & [0.89,..,0.99] & [0.89,..,0.99] & [0.89,..,0.99] & [0.89,..,0.99]  \\
       Learning rate  & [1e-4,..,1e-2] & [1e-3,..,0.2] & [1e-3,..,0.2] & [1e-3,..,0.2] & [1e-3,..,0.2] & [1e-3,..,0.2] \\
       Train frequency  & [1,..,6] & - & - & - & - & - \\
       Batch size  & [32,..64] & - & - & [32,..64] & [32,..64] & [32,..64] \\
       n epochs & - & [5,..,100] & - & - & - & - \\
       Normalize advantage & - & - & [true, false] & - & - & - \\
       \bottomrule
    \end{tabular}
    }
    \caption{Search range for hyperparameters tuning.}
    \label{tab:searchrange}
\end{table}
\begin{table}[ht!]
    \centering
    \resizebox{\columnwidth}{!}{
    \begin{tabular}{ccccc}
    \toprule
    Features & Units & Description & Min & Max\\
    \midrule
       $D$  &  kWh  & Home demand / load profile & -20 & 1000\\
       $P_{\text{pv}}$  & kW & PV generation & -1000 & 1000  \\ 
       $P_{\text{battery}}$  & kW &Battery power & -1000 & 1000  \\ 
       $SoC$  & \% & State of charge & 0 & 100  \\ 
       $price$  & \euro / kWh & Price of electricity & -1.0 & 20  \\ 
       $Day$  & numeric & Day of the month & 1 & 31  \\ 
       $Month$  & numeric & Month of the year & 1 & 12  \\ 
       $Year$  & numeric & Year & - & -  \\ 
       $h_t$  & \% & Humidity in the environment & 0 & 100  \\ 
       $T_t$  & $\degree C$ & Temperature at time $t$ & -20 & 50  \\ 
       $Rainfall_t$  & $mm/h$ & Rainfall at time $t$ & 0 & 30  \\ 
       $Windspeed_t$  & $m/s$ & Windspeed at time $t$ & 0 & 30  \\ 
       $\textit{Air pressure}_t$  & $hPa$ & Air pressure at time $t$ & 950 & 1025  \\ 
       $\textit{Is holiday}_t$  & $binary$ & If it is holiday or working day $t$ & 0 & 1  \\ 
       $\textit{Price forecast}_{t+1,..,t+4}$  & \euro / kWh & Price forecast for the next 1-hour & -1.0 & 20 \\ 
       $\textit{Demand forecast}_{t+1,..,t+4}$  & kWh & Demand forecast for the next 1-hour & -20 & 1000  \\ 
       \bottomrule
    \end{tabular}
    }    
    \caption{State space description along with their respective units and the range of values used in this work.}
    \label{tab:statespace}
\end{table}

    \item \textbf{Action Space}: The action space is defined as the charging/discharging current of the battery. Based on observed operating conditions, the current range is set to $[-60, +60]$~A, where negative values correspond to discharging and positive values to charging. For DQN, the action space is discrete - having thirteen distinct actions- but for the other algorithms, we have used a continuous action space. The bounds of the action space are derived from previous recordings within the empirical dataset to enforce the agent to take plausible actions. It is important to note that any action that overcharges or over-consumes the battery beyond the battery constraints would be considered as impossible action.
    \item \textbf{Reward Function}: The agent is trained to maximize economic profit while ensuring safe battery operation. 
The reward incorporates:
\begin{itemize}
    \item[-] \textbf{Battery operations:} include penalties for invalid actions. When the agent selects a charging or discharging action that would exceed the battery’s upper or lower capacity limits respectively, the environment replaces this action with an idle operation and assigns an additional penalty.
    \item[-] \textbf{Net demand balance:}  
    \begin{equation}
        P_{\text{net demand}} = P_{\text{pv}} - P_{\text{agent battery}} - P_{\text{demand}}
    \end{equation}

    Surplus energy is sold to the grid at a discounted rate, while deficits require purchasing electricity. It is important to note that $P_{battery}$ is different from $P_{\text{agent battery}}$ as former represents the actual battery power taken from historic data whereas later represents the battery power influenced by the RL agent.
    \item[-] \textbf{Grid dependence:} additional penalties are applied for electricity purchased from the grid. Costs are calculated as $C_{grid} = P_{grid} \cdot price$ and $C_{net}= P_{\text{net demand}}\cdot price$.
\end{itemize}

Formally, the reward is expressed as:
\begin{equation}
        R_{\text{t}} = \omega_1 \cdot C_{\text{grid}} - \omega_2 \cdot C_{\text{net}}
\label{eq:reward}
\end{equation}
where $C_{\text{grid}}$ and $C_{\text{net}}$ represent the costs associated with grid interaction and net demand. If any unsafe action is taken, the environment penalizes the agent $P_{\text{penalty}}$. 
\begin{equation}
R_t =
\begin{cases}
P_{\text{penalty}}, 
& \text{if } \mathrm{action}_t \text{ is unsafe} \\[1pt]

\omega_1 \cdot C_{\text{grid}} 
- \omega_2 \cdot C_{\text{net}}, 
& \text{if } \mathrm{action}_t \text{ is safe}
\end{cases}
\end{equation}
The weighting factors $(\omega_1, \omega_2)$ control the trade-off between economic and operational objectives.
\end{itemize}
In this work, we keep the weights identical for simplicity. Since the agent can not influence $C_{grid}$, the aim is to minimize $ C_{net}$ expression in Eq.~\eqref{eq:reward} by taking effective actions. Consequently, the overall profit calculated from Eq.~\eqref{eq:reward} is increased.
%\textcolor{revised}{In equation~\eqref{eq:reward}, two weights are used to normalize the reward and avoid the drastic bigger values. In this work, we keep the weights same because it avoids the biasness in the environment. If $\omega_1 > \omega_2$, then the weighted $C_{grid}$ is higher, meaning we impose the agent to always get positive rewards even if the actions are not meaningful. Similarly, when we have $\omega_1 < \omega_2$, the impact of $C_{grid}$ would be lowered in the reward function, which is technically incorrect.}

\subsubsection{Rule-based System}
In this system, we harness the rules from the empirical dataset and extract when specific actions (charging, discharging or idle) are taken. We have the \textbf{battery charge} information and we calculate the action in terms of current (as discussed in action space) by dividing it by the voltage. In this case, we keep the voltage as 50 V:
\begin{equation}
    A_{\text{actual action}} = P_{\text{battery}}/50
\end{equation}

%%specific references for these methods

%% shap is well used so we also tried, but it didnt work see appendix
\subsection{Policy Interpretations}
\label{section:xai-mr}
The policies obtained from the RL agents are interpreted using several explainability methods, including decision trees, and feature removal analysis \cite{Lundberg2017}. While XAI for power systems often utilizes SHAP for generating feature attributions \cite{Machlev2022}, this approach is not suitable for RL due to unstable and inconsistent attributions. Specifically,  RL policies depend on state-action trajectories rather than independent feature contributions, making additive attribution methods yield unstable explanations. Therefore, we adopt alternate explanation techniques, namely, policy distillation via decision-tree approximation, which provides structural rule extraction aligned with policy behavior \cite{Craven1996}. 
%We extracted interpretable rules from decision trees trained to approximate each RL policy, following common practice in model distillation \cite{Craven1996}. 
In this work, we ensured that the actions suggested by the agent are transparent because they can provide insights to operational actions, which can consequently be helpful to debug and validate the system. These actions are transparent since they allow the verification of SoC safety constraints and price-driven action before real-world deployment. These extracted rules were then compared with the baseline rules derived from the empirical dataset. In the feature removal analysis, we modified the environment by excluding selected state features during the training phase and evaluated the resulting change in rewards to assess the relative importance of each feature. The details are mentioned in section \ref{subsection:explain}.

\subsection{Synthetic Dataset of Building Energy Management}
\label{subsec:energyplus}
EnergyPlus is an energy systems simulator that helps in modeling the energy consumption of the buildings and is developed by the U.S. Department of Energy \cite{osti_1395882}. In this work, building energy behavior and distributed energy resource operation are simulated using the EnergyPlus example model \texttt{{Generator\_PVWatts}}, which incorporates the PVWatts photovoltaic generation model to estimate solar power output based on solar irradiance, system configuration, and weather conditions. By running the simulation under realistic weather inputs and operational settings, time-series data, including building electricity demand, photovoltaic generation, and grid electricity exchange, are extracted to construct the synthetic dataset used for training and evaluation of the proposed control framework. Unlike the empirical dataset, which is obtained from real-world measurements, the synthetic dataset is generated through controlled simulations. The dataset has a one-hour temporal resolution and spans a full year, capturing seasonal variations in weather conditions, building demand, and renewable generation.
\begin{table}[t!]
    \centering
    \resizebox{\columnwidth}{!}{
    \begin{tabular}{ccccc} \toprule
    {Features} & {A2C} & {PPO} & {SAC} & {DQN}\\ \midrule
    All  & 568.49 & 222.94 & -131.87 & -20.74 \\
    Without \textit{SoC}  & \textbf{-294.26}  & \textit{-368} & 8.53 & \textit{-61.01}  \\
    Without \textit{PV}  &  356.62 & \underline{62.38} & -4.15 & \textbf{-222.31}\\
    Without \textit{Demand}  &  591 & 436 & 151.60  & \underline{-50} \\ 
    Without \textit{Price}  & \textit{31.03}  & 310  &  \underline{-430} & 31.03 \\
    Without \textit{Demand forecast}  & 429.92 & \textbf{-9237.12} & -6.89 & -5.27 \\
    Without \textit{Price forecast}  & 332 & 212.28  & \textbf{-1573} & -6.89 \\ Without \textit{Humidity}  & \underline{32.5} & 342.13  & \textit{-824.23} & 66.12 \\ \bottomrule
\end{tabular}
}
    \caption{Comparison of cumulative rewards obtained by different reinforcement learning agents after removing individual features from the environment during the learning phase. The results illustrate the sensitivity of each agent to specific input features such as state of charge (SoC), photovoltaic (PV) generation, humidity, and forecast variables.}
    \label{tab:feature-removal}
\end{table}
\section{Results and Discussions}
\label{section-results}
In this section, we report the performance of the best-performing and optimal policies obtained from the RL and rule-based controllers. We then interpret these policies using multiple XAI techniques, enabling a systematic comparison between the explanations derived from the empirical controller and those obtained from the learned RL models. We start with a detailed description of our new empirical data set and close with a shorter treatment of the synthetic data set.

\subsection{Performance of RL Algorithms based on Rewards}
The trained agents behave differently and learned different policies. 
Some can find a better policy than an agent that does not use electrical storage, and exhibit good performance after training for about 100k steps on the training data. 
% All the models are trained on both variations of the state space, i.e., one with forecasts and one without forecasts. 
% It must be noted that a higher profit or cumulative reward is desirable. 
% We have trained different RL models, and 
Different RL agents conclude with different rewards as shown in Fig.\ref{fig:cummreward-a2c}.

\begin{figure}[ht!]
\resizebox{1.0\linewidth}{!}{
\begin{tikzpicture}
  % 1. The Decision Tree
  \node (tree) {
    \begin{forest}
      soc/.style = {ellipse, fill=greenMain!60},
      battery/.style    = {ellipse, fill=forestgreen!70},
      price/.style      = {ellipse, fill=sky!40},
      price30/.style    = {ellipse, fill=sky!55},
      price45/.style    = {ellipse, fill=sky!70},
      price60/.style    = {ellipse, fill=deepwater!60},
      humidity/.style   = {ellipse, fill=sage!60}, 
      pressure/.style   = {ellipse, fill=sand!70}, 
      pv/.style         = {ellipse, fill=sun!70},
      demand/.style     = {ellipse, fill=terracotta!50},
    demand15/.style   = {ellipse, fill=terracotta!65},
    demand30/.style   = {ellipse, fill=clay!75},
    demand60/.style   = {ellipse, fill=clay!85},
    windspeed/.style  = {ellipse, fill=wind!70},
      for tree={
        parent anchor=south,
        child anchor=north,
        l sep=1.2cm,
        s sep=0.8cm,
        edge={-Stealth, color=black, thin},
        align=center
      }
    [SoC, soc
      [D, action=green!80!black, edge label={node[midway, left, xshift=-3pt] {\small $\le 0.9$}}]
      [Battery, battery, edge label={node[midway, right, xshift=3pt] {\small $> 0.9$}}
        [Price (30m), price30, edge label={node[midway, left] {\small $\le 2952$ kW}}
          [Price, price, edge label={node[midway, left] {\small $\le -0.019$~\euro{}}}
            [D, action=green!80!black, edge label={node[midway, left] {\small $\le 0.008$}}]
            [C, action=green!80!white, edge label={node[midway, right] {\small $> 0.008$}}]
          ]
          [Humidity, humidity, edge label={node[midway, right, xshift=2pt] {\small $> -0.019$~\euro{}}}
            [Humidity, humidity, edge label={node[midway, left] {\small $\le 55.57\%$}}
              [C, action=green!80!white, edge label={node[midway, left] {\small $\le 55.43\%$}}]
              [D, action=green!80!black, edge label={node[midway, right] {\small $> 55.43\%$}}]
            ]
            [C, action=green!80!white, edge label={node[midway, right] {\small $> 55.57\%$}}]
          ]
        ]
        [D, action=green!80!black, edge label={node[midway, right] {\small $> 2952$ kW}}]
      ]
    ]
    \end{forest}
  };
  % 2. The Consolidated Legend (relative to 'tree')
  \node[anchor=north west, xshift=1cm] at (tree.north east) (legend) {
    \sffamily\small
    \begin{tabular}{@{}ll@{}}
    \multicolumn{2}{l}{\textbf{Primary Drivers}} \\ \midrule
    \tikz\node[legend box, fill=greenMain!60]{}; & SoC (Ratio) \\
    \tikz\node[legend box, fill=forestgreen!70]{};   & Battery (kW) \\
    \tikz\node[legend box, fill=sky!55]{};  & Price (30m forecast) \\
    \tikz\node[legend box, fill=sky!40]{};  & Price (Current) \\
    \tikz\node[legend box, fill=sage!60]{}; & Humidity (\%) \\
    \addlinespace[1em]
    \multicolumn{2}{l}{\textbf{Other Tested Variables}} \\ \midrule
    \tikz\node[legend box, fill=gray!10]{};  & \textcolor{gray!60}{PV Generation} \\
    \tikz\node[legend box, fill=gray!10]{};  & \textcolor{gray!60}{Hour of Day} \\
    \tikz\node[legend box, fill=gray!10]{};  & \textcolor{gray!60}{System Demand} \\
    \tikz\node[legend box, fill=gray!10]{};  & \textcolor{gray!60}{Pressure} \\
    \tikz\node[legend box, fill=gray!10]{};  & \textcolor{gray!60}{Windspeed} \\
    \bottomrule
    \end{tabular}
  };

  % 3. Action Key (at bottom of legend)
  \node[anchor=north west, yshift=-0.5cm] at (legend.south west) {
      \sffamily\small
      \begin{tabular}{@{}ll@{}}
      % \multicolumn{2}{l}{\textbf{Action}} \\ \midrule
      \tikz\node[legend box, fill=green!80!white]{}; & \textbf{C}: Charge \\
      \tikz\node[legend box, fill=green!80!black]{}; & \textbf{D}: Discharge \\
      \end{tabular}
  };

\end{tikzpicture}
}
\caption{Interpretable decision tree derived from the A2C agent’s policy. Each node represents a decision rule based on the system’s state variables, and the leaf nodes indicate whether the agent chooses to charge (C) or discharge (D) the battery. The tree provides insight into how the A2C model learns to balance charging and discharging decisions in response to changing energy conditions.}
\label{fig:a2c-dt}  
\end{figure}
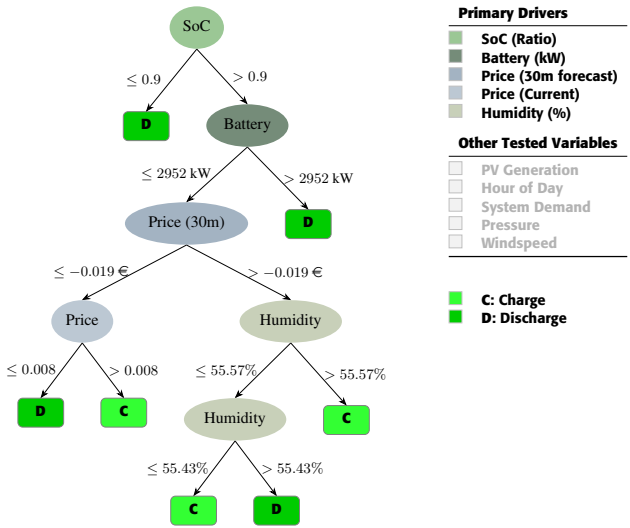

% %%%%%%Example of DT tikz fig
% \begin{figure}[ht!]
% \centering
% % \resizebox{1.0\linewidth}{!}{
% \include{a2c dt}
% % }
% \caption{Interpretable decision tree for A2C policy. Decisions are based on physical units (Wh, \%, \euro{}). Variables in gray were included in the state space but did not result in primary splits.}
% \label{fig:a2c-dt} % leave one label
% \end{figure}

%%%%%%
 
% minipage to put fig 3-4 and the legend (table) together, or just using plotted figs instead of forest

It is visible from the reward curves that A2C performs better than other algorithms. 
The environment is complex and it requires deep exploration and needs more complex policies. 
In such environments, A2C often outperforms the others \cite{de2024comparative} as it also includes the complexities of the forecasted features like price and demand. Furthermore, A2C’s on-policy advantage estimation and entropy bonus leads to both rapid early learning and high final returns on this environment.

PPO is also favourable for complex problems, as it is also evident from the learning curve shown in Fig.\ref{fig:cummreward-a2c}. 
However, we have to run several iterations with fine-tuning of different sets of hyperparameters as PPO and A2C need careful tuning of hyperparameters \cite{de2024comparative}. It is interesting to observe that the performance of on-policy algorithms (A2C and PPO) is better than the off-policy algorithms despite the advantage of sample efficiency and replay buffers in off-policy methods. 
We speculate that this advantage of the on-policy algorithms is linked to the inclusion of short-term predictions of demand and electricity price. In such settings, replay buffers used in off-policy methods may contain trajectories generated under outdated forecast conditions. In contrast, on-policy methods update the policy using freshly collected trajectories that align with the current environment dynamics. This reduces the bias introduced by stale samples and leads to more stable policy improvement.

% We have analyzed the actions taken by different agents with the actual behavior of charge and discharge of the battery as shown in Figure \ref{fig:actionsnew}. 

%%  different RL actions are closer or different 
We analyze the actions suggested by the agents to understand their control behavior and evaluate how their decisions align with feasible battery operations (see Fig. \ref{fig:actionsnew}).
However, we observe that there are some points where we have overlaps or actions close to the actual or real-world action. If we compare the A2C actions with the baseline actions, we can deduce that A2C agent take smart actions. When the baseline actions are idle, specifically at night when the demand is low, A2C charges the battery so that the reliance should be on the battery in the morning rather than buying it from grid. Consequently, this helps in earning more profit than the baseline.

\subsection{Explanations}
\label{subsection:explain}
To understand these policies, we interpret and explain them using XAI methods. 
These interpretations could be helpful for a layperson interacting with the battery to understand the policy but also for developers when optimizing the algorithms. 
We use different methods to explain the policies and compare the different levels of explanations, namely, decision trees as simplified control models and feature removal. In this work, interpretations are assessed in terms of domain knowledge and structural transparency. Human-subject evaluation is beyond the scope of this work.

\noindent\textit{Explanations with Decision Trees}

Decision trees are a valuable tool for explaining learned policies, as they are human-readable and easy to interpret. The rules extracted from these trees provide concise insights into the agent’s decision patterns; however, they should be understood as approximations of the underlying, more complex control strategy learned by the RL algorithm.
We train the decision trees after training the RL agents. 
For the sake of simplicity in training the decision trees, we generalize the agent's actions into three distinct classes: discharge, idle, and charge, and  train the trees up to a depth of 5. 

Our observation shows that A2C and PPO policies consider \textit{SoC} as a major feature to trigger the chain of following actions. % consider. 
If the \textit{SoC} exceeds 90\%, energy is drained from battery (discharge), otherwise, other features are evaluated. 
The major features that are important for the optimal policy (A2C) are: \textit{SoC, power of the battery, current price, price forecast in the next 30 minutes, and the humidity}  (see Fig.~\ref{fig:a2c-dt}). These features reflect the operational and economic factors governing the energy management problem. In particular, \textit{SoC} determines the available storage flexibility and prevents overcharging, while electricity price and short-term price forecasts guide economically optimal charge–discharge decisions. The inclusion of renewable generation and demand-related variables allows the agent to anticipate future energy availability and consumption. Consequently, the learned policy aligns with intuitive control principles of energy arbitrage and storage management, where the battery is discharged when sufficiently charged and when economic conditions are favorable.

% \begin{figure}[!h]
%     \centering
%     \includesvg[width=1.0\linewidth, keepaspectratio]{ppo dt left(1)}
%     \caption{Decision Tree (left side from root node) obtained from PPO policy}
%     \label{fig:ppo-dt}
% \end{figure}
% \begin{figure*}[!ht]
%     \centering
%     \hspace*{-1cm}
%     \resizebox{1.0\linewidth}{!}{%
%         \includesvg[width=1.5\linewidth, keepaspectratio]{ppo dt right(2)}
%     }
%     \caption{Decision Tree (right side from root node) obtained from PPO policy}
% \end{figure*}

\begin{figure}[t]
\resizebox{1.0\linewidth}{!}{
\begin{tikzpicture}
  % 1. The Decision Tree
  \node (tree) {
    \begin{forest}
      soc/.style = {ellipse, fill=greenMain!60},
      battery/.style    = {ellipse, fill=forestgreen!70},
      price/.style      = {ellipse, fill=sky!40},
      price30/.style    = {ellipse, fill=sky!55},
      price45/.style    = {ellipse, fill=sky!70},
      price60/.style    = {ellipse, fill=deepwater!60},
      humidity/.style   = {ellipse, fill=sage!60}, 
      pressure/.style   = {ellipse, fill=sand!70}, 
      pv/.style         = {ellipse, fill=sun!70},
      demand/.style     = {ellipse, fill=terracotta!50},
    demand15/.style   = {ellipse, fill=terracotta!65},
    demand30/.style   = {ellipse, fill=clay!75},
    demand60/.style   = {ellipse, fill=clay!85},
    windspeed/.style  = {ellipse, fill=wind!70},    
      for tree={
        parent anchor=south,
        child anchor=north,
        l sep=1.2cm,
        s sep=0.8cm,
        edge={-Stealth, color=black, thin},
        align=center
      }
    [SoC, soc
      [D, action=green!80!black, edge label={node[midway, left, xshift=-3pt] {\small $\le 0.9$}}]
      [PV, pv, edge label={node[midway, right, xshift=3pt] {\small $> 0.9$}}
        [Price (30m), price30, edge label={node[midway, left] {\small $\le 6957$ kW}}
          [Demand, demand, edge label={node[midway, left] {\small $\le -0.019$~\euro{}}}
            [D, action=green!80!black, edge label={node[midway, left] {\small $\le 917$}}]
            [C, action=green!80!white, edge label={node[midway, right] {\small $> 917$}}]
          ]
          [Humidity, humidity, edge label={node[midway, right, xshift=2pt] {\small $> -0.019$~\euro{}}}
            [Humidity, humidity, edge label={node[midway, left] {\small $\le 55.57\%$}}
                [Price, price, edge label={node[midway, left] {\small $\le 55.43\%$}}
                    [C, action=green!80!white, edge label={node[midway, left] {\small $\leq 0.05\%$}}]
                    [D, action=green!80!black, edge label={node[midway, right] {\small $> 0.05\%$}}]
                ]
                [D, action=green!80!black, edge label={node[midway, right] {\small $> 55.43\%$}}]
            ]
            [C, action=green!80!white, edge label={node[midway, right] {\small $> 55.57\%$}}]
          ]
        ]
        [D, action=green!80!black, edge label={node[midway, right] {\small $> 2952$ kW}}]
      ]
    ]
    \end{forest}
  };
  % 2. The Consolidated Legend (relative to 'tree')
  \node[anchor=north west, xshift=1cm] at (tree.north east) (legend) {
    \sffamily\small
    \begin{tabular}{@{}ll@{}}
    \multicolumn{2}{l}{\textbf{Primary Drivers}} \\ \midrule
    \tikz\node[legend box, fill=greenMain!60]{}; & SoC (Ratio) \\
    \tikz\node[legend box, fill=sun!70]{};   & PV (kW) \\
    \tikz\node[legend box, fill=sky!55]{};  & Price (30m forecast) \\
    \tikz\node[legend box, fill=sky!40]{};  & Price (Current) \\
    \tikz\node[legend box, fill=sage!60]{}; & Humidity (\%) \\
    \tikz\node[legend box, fill=terracotta!50]{}; & Demand (kW) \\
    \addlinespace[1em]
    \multicolumn{2}{l}{\textbf{Other Tested Variables}} \\ \midrule
    \tikz\node[legend box, fill=gray!10]{};  & \textcolor{gray!60}{Battery} \\
    \tikz\node[legend box, fill=gray!10]{};  & \textcolor{gray!60}{Hour of Day} \\
    \tikz\node[legend box, fill=gray!10]{};  & \textcolor{gray!60}{Temperature} \\
    \tikz\node[legend box, fill=gray!10]{};  & \textcolor{gray!60}{Pressure} \\
    \tikz\node[legend box, fill=gray!10]{};  & \textcolor{gray!60}{Windspeed} \\
    \bottomrule
    \end{tabular}
  };

  % 3. Action Key (at bottom of legend)
  \node[anchor=north west, yshift=-0.5cm] at (legend.south west) {
      \sffamily\small
      \begin{tabular}{@{}ll@{}}
      % \multicolumn{2}{l}{\textbf{Action}} \\ \midrule
      \tikz\node[legend box, fill=green!80!white]{}; & \textbf{C}: Charge \\
      \tikz\node[legend box, fill=green!80!black]{}; & \textbf{D}: Discharge \\
      \end{tabular}
  };

\end{tikzpicture}
}
\caption{Decision tree generated from empirical data without reinforcement learning. The nodes represent decision thresholds based on measured system variables, while the leaf nodes correspond to the charging (C) and discharging (D) actions of the battery. This data-driven model provides a baseline for understanding the inherent decision logic derived directly from observed energy system behavior.}
\label{fig:real dt}
\end{figure}
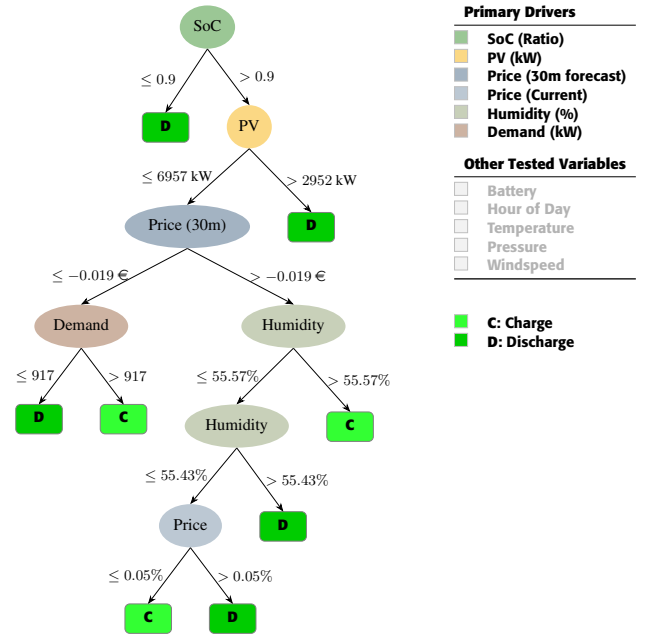

We also obtain the rules from the decision tree without the inclusion of RL agent as shown in Fig.\ref{fig:real dt}. 
It is interesting to observe that the tree has only two distinct classes: discharge and charge. 
It can be concluded that during this time frame the idle action is never taken. 
The policies that we obtain from A2C and PPO are more realistic as they consider all three actions. 
Furthermore, the features of the RL agents are also include forecasted values of demand and price into account which are ignored by our simplified decision tree. 
The anticipated values of these features likely help A2C and PPO to achieve their superior performance.  %formulate the optimal policies. 

Finally, we note that these decision trees cannot replace the DRL agents. Both the extracted trees from the existing control strategy (Fig. \ref{fig:real dt}) or approximating the RL model (Fig. \ref{fig:a2c-dt}) perform poorer than the full RL models. We still need RL policy as it uses the forecasted values to take more complex actions. For example,  it learns anticipatory charging and price-driven arbitrage, which are absent in the baseline environment.
If a simple transparent system is desired, the decision tree of the RL model  (Fig.\ref{fig:a2c-dt}) could be used instead of the existing policy, as it outperforms the current one while being fully transparent.
%\textcolor{revised}{It is important to consider that the decision trees trained on empirical data reproduce the behaviour of the existing environment, whereas the DRL agents learn a new control strategy that significantly increases the cumulative reward as shown in Table~\ref{tab:comparison-dts-with-rl}. The trees extracted from the RL policy therefore, explain an optimized policy rather than replacing the learning process. We still need RL policy as it uses the forecasted values to take intelligent actions. Moreover, the RL agent does not replicate the baseline policy; it learns anticipatory charging and price-driven arbitrage, which are absent in the baseline environment.}

\begin{table}[th!]
    \centering
    \begin{tabular}{cc} \toprule
    {Models} & {Rewards/Profit gains} \\ \midrule
    A2C  & \textbf{568.49} \\
    Baseline &  0 \\
    DT - A2C  & \textit{29.34} \\
    DT - Baseline &  -1420.42  \\ \midrule
    PPO (EnergyPlus) & 412 \\
    A2C (EnergyPlus) & 400 \\
    \bottomrule
\end{tabular}
    \caption{Comparison of cumulative rewards for the best-performing RL agent (A2C), the baseline controller, and the corresponding decision-tree (DT) approximations along with the approximations of best performing model PPO (EnergyPlus) on simulated data. The RL agent outperforms both the baseline and the tree-based approximations of its policy.}
    \label{tab:comparison-dts-with-rl}
\end{table}

\noindent\textit{Explanations with Feature Removal}

Complementary to decision trees, we also explore the feature removal methodology to find out the impact of excluding features on the cumulative rewards of all algorithms. 
 Instead of testing all features, we conduct experiments on the features that we believe would have the most significant impact, including \textit{pv, demand, price, SoC, humidity, demand forecast and price forecast}, partially guided by our results from the decision trees.
% \begin{table}[h]
%     \centering
%     \begin{tabular}{cccc} \toprule
%     {Algorithms} & {With all features} & {Without $PV$} & {Without $Demand$} \\ \midrule
%     DQN  & -20.739293 & -222.310529 & -50   \\
%     PPO  & 222.939232  & 62.375794 & 436    \\
%     A2C  &  568.490027 & 356.624536 & 591 \\
%     SAC  &  -131.867470 & -4.145078 & 151.596301   \\ 
%     DDPG  & -25  & -25  &  -25  \\
%     TD3  &  -25 & -25 &  -25   \\\midrule
%        & {Without $Price$} & {Without $Demand Forecast$} & {Without $Price forecast$}\\\midrule
%      DQN  & 31.032414 & -5.27& -6.89   \\
%     PPO  & 310 & 100.53 &  212.28  \\
%     A2C  & 31.032414&429.922636 & 332   \\
%     SAC  & -430 & -923711.2& -1573  \\ 
%     DDPG  & -25  & -25  &  -25  \\
%     TD3  &  -25 & -25 &  -25   \\ \\\midrule
%        & {Without $SoC$} &  & \\\midrule
%      DQN  & 549.83 & -5.27& -6.89   \\
%     PPO  & -368 & 100.53 &  212.28  \\
%     A2C  & -294.26&429.922636 & 332   \\
%     SAC  &  & -923711.2& -1573  \\ 
%     DDPG  &   & -25  &  -25  \\
%     TD3  &  -25 & -25 &  -25   \\ \bottomrule
% \end{tabular}
%     \caption{Cumulative rewards of RL policies after removing the features from the environment}
%     \label{tab:feature-removal}
% \end{table}

Table \ref{tab:feature-removal} depicts the impact on cumulative rewards or profit gain after dropping the feature from the state space. 
From these experiments, we can conclude that every feature plays a different role in formulating the different RL policies. 
We observe that all the features are important for these strategies. 
If we drop the \textit{PV generation} feature from the environment, the profit gain is dropped for on-policy RL algorithms. 
Similarly, \textit{price} is a key feature for these algorithms. 
However, it is interesting to observe that \textit{demand} is not important for on-policy algorithms (A2C, PPO) rather future values of demand and price (\textit{demand forecast, price forecast}) have an evident importance. This leads to the fact that \textit{demand} is not important because it should be fulfilled in any case. However based on forecasted demand, the agent can take action on battery to prepare it for future.
This importance is also depicted by the decision tree (Fig. \ref{fig:ppo-dt} in the Appendix). 
Hence, the explanations are consistent with the corresponding decision-tree representations. Nevertheless, these trees constitute simplified approximations of the underlying policy and cannot capture all relevant decision-making nuances. The RL model retains a more comprehensive understanding of the system dynamics, which is only partially reflected in the tree-based approximations (see Table \ref{tab:comparison-dts-with-rl}).

\subsection{Comparing Policy Performance on a Synthetic Dataset}
\label{subsec:policycons}
The generalization capability of the framework is evaluated by testing the trained policies on unseen datasets. This step helps assess the robustness of the learned policies and reduces the risk of overfitting to a single dataset. For a fair comparison, the trained policies trained on the empirical dataset were evaluated on the simulated dataset.

To complement the training on the empirical data set, we generate a synthetic dataset using EnergyPlus, which simulates building energy consumption, power generation from PV, along with corresponding weather conditions \cite{osti_1395882}. The simulated data covers a temporal span of one year, resulting in more than 8,500 records. This dataset captures seasonal variations that influence PV generation and building load profiles for buildings located across different geographical sites.

\begin{figure*}[ht]
    \centering
    \includesvg[width=\linewidth]{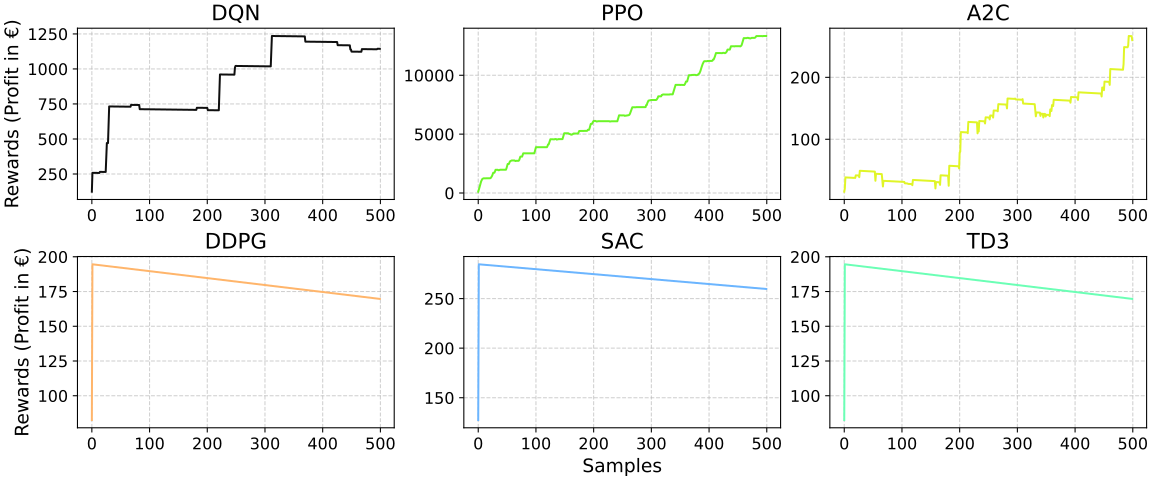}
    \caption{Cumulative rewards on synthetic dataset proves that on-policy algorithms (PPO and A2C)) are suited for synthetic dataset. However an off-policy algorithm (DQN) is applicable for building model as well.}
    \label{fig:lingraph_synthetic}
\end{figure*}

The evaluation indicates a performacne similar to the real-world building dataset (see Fig.~\ref{fig:lingraph_synthetic}). Again, the on-policy reinforcement learning agents consistently achieve higher rewards than the off-policy ones. Interestingly, the DQN agent demonstrates slightly improved performance on the synthetic dataset, suggesting that the policy learned by DQN might be generally usable even if it is not always among the best performers.
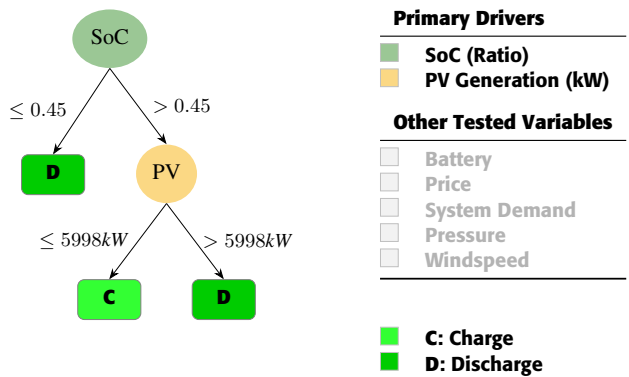
\begin{figure}[th!]
\resizebox{1.0\linewidth}{!}{
\begin{tikzpicture}
  % 1. The Decision Tree
  \node (tree) {
    \begin{forest}
    soc/.style = {ellipse, fill=greenMain!60},
      battery/.style    = {ellipse, fill=forestgreen!70},
      price/.style      = {ellipse, fill=sky!40},
      price30/.style    = {ellipse, fill=sky!55},
      price45/.style    = {ellipse, fill=sky!70},
      price60/.style    = {ellipse, fill=deepwater!60},
      humidity/.style   = {ellipse, fill=sage!60}, 
      pressure/.style   = {ellipse, fill=sand!70}, 
      pv/.style         = {ellipse, fill=sun!70},
      demand/.style     = {ellipse, fill=terracotta!50},
    demand15/.style   = {ellipse, fill=terracotta!65},
    demand30/.style   = {ellipse, fill=clay!75},
    demand60/.style   = {ellipse, fill=clay!85},
    windspeed/.style  = {ellipse, fill=wind!70},
      for tree={
        parent anchor=south,
        child anchor=north,
        l sep=1.2cm,
        s sep=0.8cm,
        edge={-Stealth, color=black, thin},
        align=center
      }
    [SoC, soc
      [D, action=green!80!black, edge label={node[midway, left, xshift=-3pt] {\small $\le 0.45$}}]
      [PV, pv, edge label={node[midway, right, xshift=3pt] {\small $> 0.45$}}
        [C, action=green!80!white, edge label={node[midway, left] {\small $\le 5998 kW$}}]
        [D, action=green!80!black, edge label={node[midway, right] {\small $> 5998 kW$}}]
      ]
    ]
    \end{forest}
  };
  % 2. The Consolidated Legend (relative to 'tree')
  \node[anchor=north west, xshift=1cm] at (tree.north east) (legend) {
    \sffamily\small
    \begin{tabular}{@{}ll@{}}
    \multicolumn{2}{l}{\textbf{Primary Drivers}} \\ \midrule
    \tikz\node[legend box, fill=greenMain!60]{}; & SoC (Ratio) \\
    \tikz\node[legend box, fill=sun!70]{};   & PV Generation (kW) \\
    \addlinespace[1em]
    \multicolumn{2}{l}{\textbf{Other Tested Variables}} \\ \midrule
    \tikz\node[legend box, fill=gray!10]{};  & \textcolor{gray!60}{Battery} \\
    \tikz\node[legend box, fill=gray!10]{};  & \textcolor{gray!60}{Price} \\
    \tikz\node[legend box, fill=gray!10]{};  & \textcolor{gray!60}{System Demand} \\
    \tikz\node[legend box, fill=gray!10]{};  & \textcolor{gray!60}{Pressure} \\
    \tikz\node[legend box, fill=gray!10]{};  & \textcolor{gray!60}{Windspeed} \\
    \bottomrule
    \end{tabular}
  };

  % 3. Action Key (at bottom of legend)
  \node[anchor=north west, yshift=-0.5cm] at (legend.south west) {
      \sffamily\small
      \begin{tabular}{@{}ll@{}}
      % \multicolumn{2}{l}{\textbf{Action}} \\ \midrule
      \tikz\node[legend box, fill=green!80!white]{}; & \textbf{C}: Charge \\
      \tikz\node[legend box, fill=green!80!black]{}; & \textbf{D}: Discharge \\
      \end{tabular}
  };

\end{tikzpicture}% \begin{minipage}{0.45\textwidth}
}
\caption{DT derived from A2C agent’s policy using simulated dataset.}
\label{fig:a2c-dt-synthetic} 
\end{figure}

Again, we also interpret the actions taken by these RL models. The $SoC$ feature remains important for all agents, even when the policies are evaluated on different datasets. This observation is consistent with the interpretations obtained from the previous experiments. However, the approximation obtained from the A2C agent is relatively simple, as it primarily relies on only two features: $SoC$ and $PV$ (see Fig~\ref{fig:a2c-dt-synthetic}).

The approximation derived from the PPO agent is very similar to the one obtained from the real-world dataset (see Fig~\ref{fig:ppo-dt-synthetic} in the Appendix). Note that the RL policies themselves are not retrained; rather, the decision trees are fitted to the state–action pairs observed when the trained policies are evaluated on the synthetic dataset. This consistency indicates that the learned policy is reliable and strengthens the evidence supporting the generalization capability of the proposed framework.

Overall, this second data set indicates that the proposed framework is easily applicable to both real-world and synthetically generated building environments, highlighting its potential application to different building structures and operational conditions.  

% \textcolor{revised}{After evaluation, we found out that all the agents are performing quite identical as they performed on the real-world building (see Fig~\ref{fig:lingraph_synthetic}). Building environments are quite suitable for the on-policy agents as these agents are efficiently performing giving best rewards. Interestingly, DQN perform better on this data which means that the policy learnt by the DQN agent is significantly suitable for the synthetic dataset. y management  the on-policy algorithms outperforms on the new building structure as well. However, DQN agent also find this synthetically generated building feasible to perform its actions . }

\section{Conclusion and Future Direction}
\label{section-conclude}
In summary, we have developed and presented an explainable reinforcement learning (XRL) framework that optimizes and interprets energy management in a building. We have demonstrated the proposed explainable DRL both in a synthetic and a novel empirical data setting. In this work, we apply different off-policy and on-policy algorithms to optimize the energy management in a single building. The real-world building data is used in this building to train RL agent and get efficient strategies. A wide space of 21 different features including price, demand, weather information  have been considered during the training phase. Experimental results have shown on-policy algorithms (A2C, PPO) are performing better than off-policy algorithms. Consequently, they provide optimal strategies for our complex system. These policies are explained with two different techniques to make it understandable for a user so that he can gain trust and rely on the actions of the agent. The techniques that we have used in this work for interpretation are decision trees and feature removal.

In conclusion, compared to the baseline controller, our approach reduces operational costs while providing transparent and interpretable decision-making, thereby increasing trust in the learned policies. We hope that this work encourages broader adoption of explainable AI methods and greater consideration of transparency in energy system algorithms. For critical infrastructure, such transparency is already required under the European Union's AI Act \cite{EU_AI_explain} and is increasingly discussed in other regions, including the United States \cite{DHS_AI}. Beyond regulatory demands, explainability offers additional benefits to AI developers and domain experts by highlighting influential features and enabling the extraction of simplified decision-tree representations of complex RL policies. Furthermore, the data-driven nature of our approach enhances its flexibility, as a policy trained for one building can be adapted to others through transfer learning. In this sense, our XRL framework complements detailed model-predictive controllers (MPC): on the one hand, XRL can offer a fully data-driven alternative that is updated through retraining rather than model redesign; on the other hand, MPC formulations can be informed or simplified using interpretable rules derived from the XRL framework.

In the future, we aim to extend the presented XRL framework beyond the building level to additional components of power system operation. Potential applications include emergency control in transmission grids \cite{L2RPN_cite}, coordinated EV charging, and congestion management or redispatch tasks, where data-driven control has shown increasing promise. Another important direction concerns aligning local control policies with broader grid-level objectives. In this study, we adopted the perspective of a single building operating under predefined buying and selling rules. However, additional reward terms could be integrated to incentivize behavior that supports system stability. While current price signals implicitly reflect some grid conditions, further incentives—potentially provided by transmission system operators (TSOs) to reduce redispatch needs or mitigate congestion—could guide local charging and discharging strategies toward globally beneficial outcomes. Such extensions would enable a more holistic integration of XRL approaches into power system operation and planning.

\section*{Acknowledgments}
We gratefully acknowledge funding from the Helmholtz Association and the Networking Fund through Helmholtz AI and under grant no. VH-NG-1727. We also acknowledge the usage of CPU resources of the HAICORE@KIT high-performance computing infrastructure. Furthermore, the authors gratefully acknowledge the computing time provided on the high-performance computer HoreKa by the National High-Performance Computing Center at KIT (NHR@KIT). This center is jointly supported by the Federal Ministry of Education and Research and the Ministry of Science, Research and the Arts of Baden-Württemberg, as part of the National High-Performance Computing (NHR) joint funding program (https://www.nhr-verein.de/en/our-partners). HoreKa is partly funded by the German Research Foundation (DFG).

\section*{Declaration of generative AI and AI-assisted technologies in the manuscript preparation process}
During the preparation of this work the authors used ChatGPT in order to rephrase the content. After using this tool, the authors reviewed and edited the content as needed and take full responsibility for the content of the published article.

\appendices
\section{\break Decision Trees for PPO models}
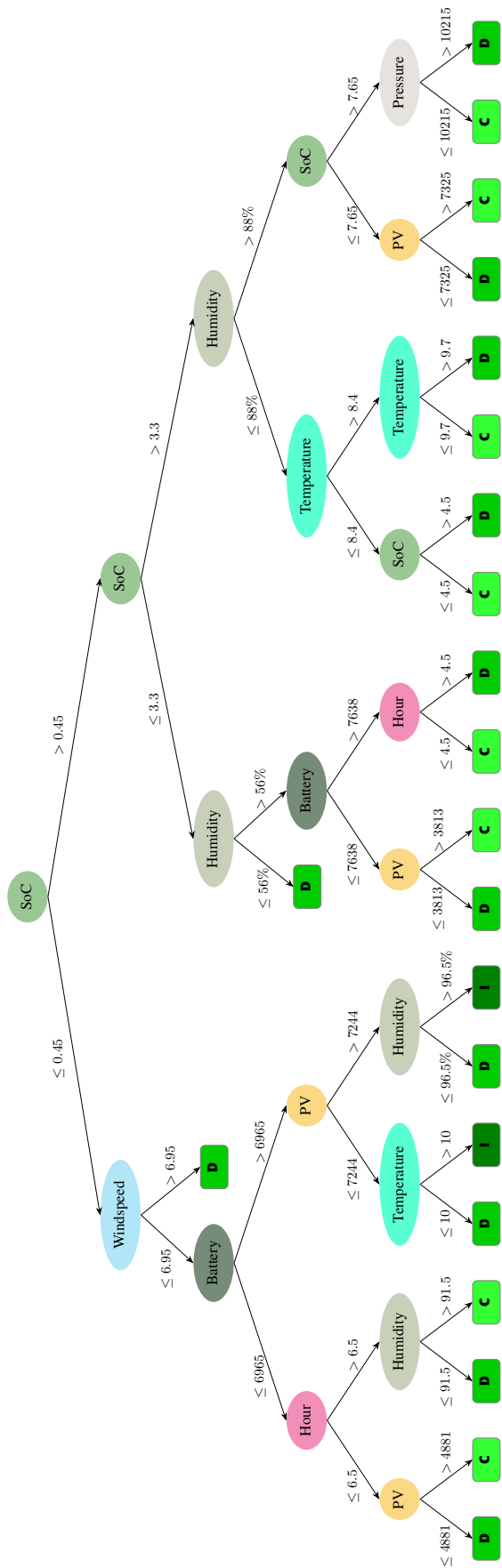
\begin{sidewaysfigure*}[ht!]
\resizebox{1.0\linewidth}{!}{
\begin{tikzpicture}
  % 1. The Decision Tree
  \node (tree) {
    \begin{forest}
      soc/.style = {ellipse, fill=greenMain!60},
      battery/.style    = {ellipse, fill=forestgreen!70},
      price/.style      = {ellipse, fill=sky!40},
      price30/.style    = {ellipse, fill=sky!55},
      price45/.style    = {ellipse, fill=sky!70},
      price60/.style    = {ellipse, fill=deepwater!60},
      humidity/.style   = {ellipse, fill=sage!60}, 
      pressure/.style   = {ellipse, fill=sand!70}, 
      pv/.style         = {ellipse, fill=sun!70},
      demand/.style     = {ellipse, fill=terracotta!50},
    demand15/.style   = {ellipse, fill=terracotta!65},
    demand30/.style   = {ellipse, fill=clay!75},
    demand60/.style   = {ellipse, fill=clay!85},
    windspeed/.style  = {ellipse, fill=wind!70},
    hour/.style ={ellipse, fill=hourMain},
    temp/.style ={ellipse, fill=temperature},
    for tree={
        parent anchor=south,
        child anchor=north,
        l sep=1.2cm,
        s sep=0.8cm,
        edge={-Stealth, color=black, thin},
        align=center
      }
    [SoC, soc
        [Windspeed, windspeed, edge label={node[midway, left, yshift=10pt, xshift=15pt] {\small $\leq 0.45$}}
            [Battery, battery, edge label={node[midway, left, xshift=-3pt] {\small $\leq 6.95$}}
                [Hour, hour,edge label={node[midway, left, xshift=-5pt] {\small $\leq 6965$}}
                    [PV, pv, edge label={node[midway, left, xshift=-3pt] {\small $\leq 6.5$}}
                        [D, action=green!80!black, edge label={node[midway, left] {\small $\leq 4881$}}]
                        [C, action=green!80!white, edge label={node[midway, right] {\small $> 4881$}}]
                    ]
                    [Humidity, humidity, edge label={node[midway, right, xshift=3pt] {\small $> 6.5$}}
                        [D, action=green!80!black, edge label={node[midway, left] {\small $\leq 91.5$}}]
                        [C, action=green!80!white, edge label={node[midway, right] {\small $> 91.5$}}]
                    ]
                ]
                [PV, pv, edge label={node[midway, right, xshift=10pt] {\small $> 6965$}}
                    [Temperature, temp,edge label={node[midway, left, xshift=-5pt] {\small $\leq 7244$}}
                        [D, action=green!80!black, edge label={node[midway, left] {\small $\leq 10$}}]
                        [I, action=green!50!black, edge label={node[midway, right] {\small $> 10$}}]
                    ]
                    [Humidity, humidity, edge label={node[midway, right, xshift=5pt] {\small $> 7244$}}
                        [D, action=green!80!black, edge label={node[midway, left] {\small $\leq 96.5\%$}}]
                        [I, action=green!50!black, edge label={node[midway, right] {\small $> 96.5\%$}}]
                    ]
                ]
            ]
            [D, action=green!80!black, edge label={node[midway, right] {\small $> 6.95$}}]
        ]
        [SoC, soc, edge label={node[midway, right, yshift=10pt, xshift=-15pt] {\small $> 0.45$}}
            [Humidity, humidity,edge label={node[midway, right, yshift=8pt, xshift=-15pt] {\small $\leq 3.3$}}
               [D, action=green!80!black, edge label={node[midway, left] {\small $\leq 56\%$}}]
                [Battery, battery, edge label={node[midway, right] {\small $> 56\%$}}
                    [PV, pv,  edge label={node[midway, left, xshift=-5pt] {\small $\leq 7638$}}
                        [D, action=green!80!black, edge label={node[midway, left, yshift=5pt, xshift=5pt] {\small $\leq 3813$}}]
                         [C, action=green!80!white, edge label={node[midway, right, yshift=5pt, xshift=-5pt] {\small $> 3813$}}]
                    ]
                    [Hour, hour,edge label={node[midway, right, xshift=3pt] {\small $> 7638$}}
                        [C, action=green!80!white, edge label={node[midway, left] {\small $\leq 4.5$}}]
                        [D, action=green!80!black, edge label={node[midway, right] {\small $> 4.5$}}]
                    ]
                ]
            ]
            [Humidity, humidity,edge label={node[midway, right, yshift=8pt, xshift=-8pt] {\small $> 3.3$}}
                [Temperature, temp, edge label={node[midway, left, yshift=5pt, xshift=5pt] {\small $\leq 88\%$}}
                    [SoC, soc, edge label={node[midway, left, xshift=-3pt] {\small $\leq 8.4$}}
                        [C, action=green!80!white, edge label={node[midway, left] {\small $\leq 4.5$}}]
                        [D, action=green!80!black, edge label={node[midway, right] {\small $> 4.5$}}]
                    ]
                    [Temperature, temp,edge label={node[midway, right, xshift=3pt] {\small $> 8.4$}}
                        [C, action=green!80!white, edge label={node[midway, left] {\small $\leq 9.7$}}]
                        [D, action=green!80!black, edge label={node[midway, right] {\small $> 9.7$}}]
                    ]
                ]
                [SoC, soc, edge label={node[midway, right, yshift=8pt, xshift=-8pt] {\small $> 88\%$}}
                    [PV, pv,edge label={node[midway, left] {\small $\leq 7.65$}}
                        [D, action=green!80!black, edge label={node[midway, left] {\small $\leq 7325$}}]
                        [C, action=green!80!white, edge label={node[midway, right] {\small $> 7325$}}]
                    ]
                    [Pressure, pressure, edge label={node[midway, right] {\small $> 7.65$}}
                        [C, action=green!80!white, edge label={node[midway, left] {\small $\leq 10215$}}]
                        [D, action=green!80!black, edge label={node[midway, right] {\small $> 10215$}}]
                    ]
                ]
            ]
        ]
      ]
    ]
    \end{forest}
  };
  % 2. The Consolidated Legend (relative to 'tree')
  % \node[anchor=north west, xshift=1cm] at (tree.north east) (legend) {
  %   \sffamily\small
  %   \begin{tabular}{@{}ll@{}}
  %   \multicolumn{2}{l}{\textbf{Primary Drivers}} \\ \midrule
  %   \tikz\node[legend box, fill=orange!40]{}; & SoC (Ratio) \\
  %   \tikz\node[legend box, fill=yellow!20]{};   & PV (kW) \\
  %   \tikz\node[legend box, fill=blue!40]{};  & Price (30m forecast) \\
  %   \tikz\node[legend box, fill=blue!20]{};  & Price (Current) \\
  %   \tikz\node[legend box, fill=purple!20]{}; & Humidity (\%) \\
  %   \tikz\node[legend box, fill=brown!20]{}; & Demand (kW) \\
  %   \addlinespace[1em]
  %   \multicolumn{2}{l}{\textbf{Other Tested Variables}} \\ \midrule
  %   \tikz\node[legend box, fill=gray!10]{};  & \textcolor{gray!60}{Battery} \\
  %   \tikz\node[legend box, fill=gray!10]{};  & \textcolor{gray!60}{Hour of Day} \\
  %   \tikz\node[legend box, fill=gray!10]{};  & \textcolor{gray!60}{Temperature} \\
  %   \tikz\node[legend box, fill=gray!10]{};  & \textcolor{gray!60}{Pressure} \\
  %   \tikz\node[legend box, fill=gray!10]{};  & \textcolor{gray!60}{Windspeed} \\
  %   \bottomrule
  %   \end{tabular}
  % };

  % % 3. Action Key (at bottom of legend)
  % \node[anchor=north west, yshift=-0.5cm] at (legend.south west) {
  %     \sffamily\small
  %     \begin{tabular}{@{}ll@{}}
  %     % \multicolumn{2}{l}{\textbf{Action}} \\ \midrule
  %     \tikz\node[legend box, fill=green!80!white]{}; & \textbf{C}: Charge \\
  %     \tikz\node[legend box, fill=green!80!black]{}; & \textbf{D}: Discharge \\
  %     \end{tabular}
  % };

\end{tikzpicture}
} 
\caption{Decision tree representation of the PPO agent’s learned policy on synthetic data. Each internal node represents a decision condition based on the system’s state variables, while the terminal leaves correspond to the battery idle (I), charging (C) and discharging (D) actions. The deeper structure of the tree reflects the PPO agent’s more complex decision-making process for energy management under varying operational conditions.}
 \label{fig:ppo-dt-synthetic}
\end{sidewaysfigure*}

We complement interpretability results, specifically the decision trees, from the main text by providing the decision tree for PPO model for the empirical data in Fig.~\ref{fig:ppo-dt} and the decision tree for PPO model for the synthetic data in Fig.~\ref{fig:ppo-dt-synthetic}.

\begin{sidewaysfigure*}[ht!]
\resizebox{1.0\linewidth}{!}{
\begin{tikzpicture}
  % 1. The Decision Tree
  \node (tree) {
    \begin{forest}
      soc/.style = {ellipse, fill=greenMain!60},
      battery/.style    = {ellipse, fill=forestgreen!70},
      price/.style      = {ellipse, fill=sky!40},
      price30/.style    = {ellipse, fill=sky!55},
      price45/.style    = {ellipse, fill=sky!70},
      price60/.style    = {ellipse, fill=deepwater!60},
      humidity/.style   = {ellipse, fill=sage!60}, 
      pressure/.style   = {ellipse, fill=sand!70}, 
      pv/.style         = {ellipse, fill=sun!70},
      demand/.style     = {ellipse, fill=terracotta!50},
    demand15/.style   = {ellipse, fill=terracotta!65},
    demand30/.style   = {ellipse, fill=clay!75},
    demand60/.style   = {ellipse, fill=clay!85},
    windspeed/.style  = {ellipse, fill=wind!70},
    for tree={
        parent anchor=south,
        child anchor=north,
        l sep=1.2cm,
        s sep=0.8cm,
        edge={-Stealth, color=black, thin},
        align=center
      }
    [SoC, soc
      [Pressure, pressure, edge label={node[midway, left, yshift=10pt, xshift=10pt] {\small $\leq 0.9$}}
        [Price (60m), price60, edge label={node[midway, left, yshift=5pt, xshift=5pt] {\small $\leq 1003.29$ hPa}}
            [Price, price, edge label={node[midway, left, xshift=-5pt] {\small $\leq 0.04$}}
                [Humidity, humidity, edge label={node[midway, left, xshift=-3pt] {\small $\leq 0.02$}}
                    [I, action=green!50!black, edge label={node[midway, left] {\small $\le 78.66\%$}}]
                    [D, action=green!80!black, edge label={node[midway, right] {\small $> 78.66\%$}}]
                ]
                [D, action=green!80!black, edge label={node[midway, right] {\small $> 0.02$}}]
            ]
            [Price (45m), price45, edge label={node[midway, right, xshift=5pt] {\small $> 0.04$}}
              [Price (60m), price60,edge label={node[midway, left, xshift=-3pt] {\small $\leq 0.202$}}
                  [Demand, demand, edge label={node[midway, left, xshift=-3pt] {\small $\leq 0.202$}}
                    [D, action=green!80!black, edge label={node[midway, left] {\small $\leq 582.69$ kW}}]
                    [I, action=green!50!black, edge label={node[midway, right] {\small $> 582.69$ kW}}]
                  ]
                  [I, action=green!50!black, edge label={node[midway, right] {\small $> 0.202$}}]
                ]
                [I, action=green!50!black, edge label={node[midway, right] {\small $> 0.202$}}]
            ]
        ]
        [Price (45m), price45, edge label={node[midway, right, yshift=5pt, xshift=-5pt] {\small $> 1003.29$ hPa}}
            [I, action=green!50!black, edge label={node[midway, left] {\small $\le 0.064$}}]
            [Humidity, humidity, edge label={node[midway, right] {\small $\> 0.064 $}}
                [I, action=green!50!black, edge label={node[midway, left] {\small $\le 73.89\%$}}]
                [D, action=green!80!black, edge label={node[midway, right] {\small $> 73.89\%$}}]
            ]
        ]
      ]
      [PV, pv, edge label={node[midway, right, yshift=10pt, xshift=-10pt] {\small $> 0.9$}}
        [Demand (60m), demand60, edge label={node[midway, left, yshift=5pt, xshift=10pt] {\small $\le 2269$ kW}}
          [Demand, demand, edge label={node[midway, left] {\small $\le-0.019$~\euro{}}}
            [D, action=green!80!black, edge label={node[midway, left] {\small $\le 917$}}]
            [C, action=green!80!white, edge label={node[midway, right] {\small $> 917$}}]
          ]
          [Humidity, humidity, edge label={node[midway, right, xshift=2pt] {\small $> -0.019$~\euro{}}}
            [Humidity, humidity, edge label={node[midway, left] {\small $\le 55.57\%$}}
                [Price, price, edge label={node[midway, left] {\small $\le 55.43\%$}}
                    [C, action=green!80!white, edge label={node[midway, left] {\small $\leq 0.05\%$}}]
                    [D, action=green!80!black, edge label={node[midway, right] {\small $> 0.05\%$}}]
                ]
                [D, action=green!80!black, edge label={node[midway, right] {\small $> 55.43\%$}}]
            ]
            [C, action=green!80!white, edge label={node[midway, right] {\small $> 55.57\%$}}]
          ]
        ]
        [SoC, soc, edge label={node[midway, right, yshift=5pt, xshift=-5pt] {\small $>2269$ kW}}
            [PV, pv, edge label={node[midway, left, yshift=5pt, xshift=5pt] {\small $\le 6.3\%$}}
                [I, action=green!50!black, edge label={node[midway, left, yshift=10pt, xshift=10pt] {\small$\le 2380$ kW}}]
                [Price (30m), price30, edge label={node[midway, right, yshift=10pt, xshift=-10pt] {\small$> 2380$ kW}}
                    [Demand (15m), demand15,  edge label={node[midway, left] {\small $\le 0.082$}}
                        [C, action=green!80!white, edge label={node[midway, left] {\small $\leq 957.52$ kW}}]
                        [D, action=green!80!black, edge label={node[midway, right] {\small $> 957.52$ kW}}]
                    ]
                    [C, action=green!80!white, edge label={node[midway, right] {\small $> 0.082$}}]
                ]
            ]
            [Humidity, humidity,edge label={node[midway, right, yshift=5pt, xshift=-5pt] {\small $> 6.3\%$}}
                [SoC, soc, edge label={node[midway, left, xshift=-5pt] {\small $\leq 66.48\%$}}
                    [Demand (30m), demand30, edge label={node[midway, left, xshift=-3pt] {\small $\leq 20.7$}}
                        [D, action=green!80!black, edge label={node[midway, left] {\small $\leq 931.09$}}]
                        [C, action=green!80!white, edge label={node[midway, right] {\small $> 931.09$}}]
                    ]
                    [C, action=green!80!white, edge label={node[midway, right, xshift=3pt] {\small $> 20.7$}}]
                ]
                [Windspeed, windspeed,  edge label={node[midway, right, xshift=5pt] {\small $> 66.48\%$}}
                    [D, action=green!80!black, edge label={node[midway, left] {\small $\leq 0.38$}}]
                    [Price (30m), price30, edge label={node[midway, right, xshift=3pt] {\small $> 0.38$}}
                        [C, action=green!80!white, edge label={node[midway, left] {\small $\leq 0.01$}}]
                        [D, action=green!80!black, edge label={node[midway, right] {\small $> 0.01$}}]
                    ]
                ]
            ]
        ]
      ]
    ]
    \end{forest}
  };
  % 2. The Consolidated Legend (relative to 'tree')
  % \node[anchor=north west, xshift=1cm] at (tree.north east) (legend) {
  %   \sffamily\small
  %   \begin{tabular}{@{}ll@{}}
  %   \multicolumn{2}{l}{\textbf{Primary Drivers}} \\ \midrule
  %   \tikz\node[legend box, fill=orange!40]{}; & SoC (Ratio) \\
  %   \tikz\node[legend box, fill=yellow!20]{};   & PV (kW) \\
  %   \tikz\node[legend box, fill=blue!40]{};  & Price (30m forecast) \\
  %   \tikz\node[legend box, fill=blue!20]{};  & Price (Current) \\
  %   \tikz\node[legend box, fill=purple!20]{}; & Humidity (\%) \\
  %   \tikz\node[legend box, fill=brown!20]{}; & Demand (kW) \\
  %   \addlinespace[1em]
  %   \multicolumn{2}{l}{\textbf{Other Tested Variables}} \\ \midrule
  %   \tikz\node[legend box, fill=gray!10]{};  & \textcolor{gray!60}{Battery} \\
  %   \tikz\node[legend box, fill=gray!10]{};  & \textcolor{gray!60}{Hour of Day} \\
  %   \tikz\node[legend box, fill=gray!10]{};  & \textcolor{gray!60}{Temperature} \\
  %   \tikz\node[legend box, fill=gray!10]{};  & \textcolor{gray!60}{Pressure} \\
  %   \tikz\node[legend box, fill=gray!10]{};  & \textcolor{gray!60}{Windspeed} \\
  %   \bottomrule
  %   \end{tabular}
  % };

  % % 3. Action Key (at bottom of legend)
  % \node[anchor=north west, yshift=-0.5cm] at (legend.south west) {
  %     \sffamily\small
  %     \begin{tabular}{@{}ll@{}}
  %     % \multicolumn{2}{l}{\textbf{Action}} \\ \midrule
  %     \tikz\node[legend box, fill=green!80!white]{}; & \textbf{C}: Charge \\
  %     \tikz\node[legend box, fill=green!80!black]{}; & \textbf{D}: Discharge \\
  %     \end{tabular}
  % };

\end{tikzpicture}
} 
\caption{Decision tree representation of the PPO agent’s learned policy. Each internal node represents a decision condition based on the system’s state variables, while the terminal leaves correspond to the battery idle (I), charging (C) and discharging (D) actions. The deeper structure of the tree reflects the PPO agent’s more complex decision-making process for energy management under varying operational conditions.}
\label{fig:ppo-dt}
\end{sidewaysfigure*}

% \bibliographystyle{unsrt} 
% \bibliography{references} 

% profile of authors
\clearpage

\bibliographystyle{unsrt} 
\bibliography{references} 

\newpage

\begin{IEEEbiography}[{\includegraphics[width=1in,height=1.25in,clip,keepaspectratio]{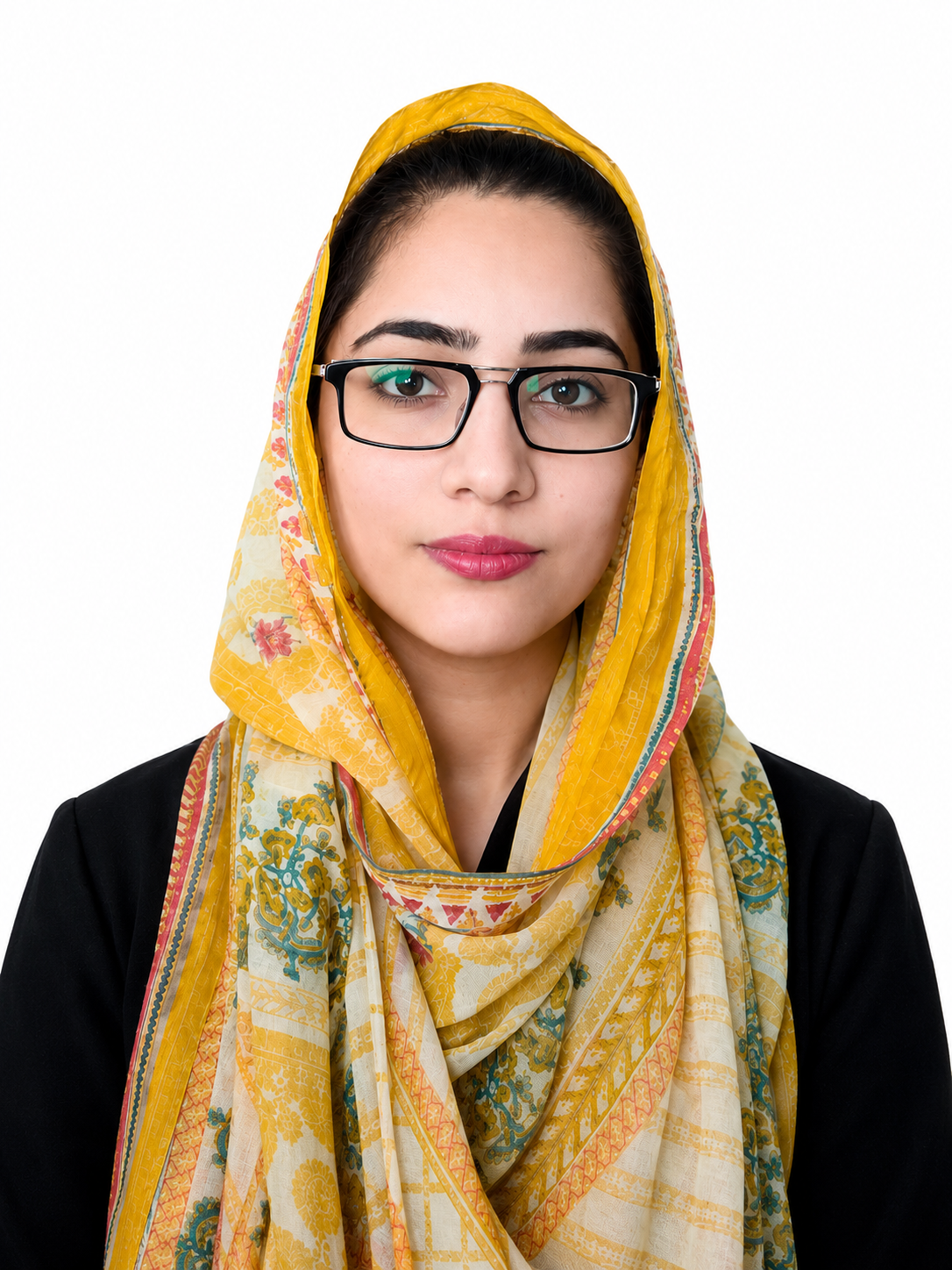}}]{Hallah Shahid Butt} received the B.Sc. degree in Software Engineering from Fatima Jinnah Women University and M.Sc. degree in Information Technology from National University of Sciences and Technology , Pakistan. She is doing Ph.D. degree from Karlsruhe Institute of Technology, Germany. Her research interests include reinforcement learning, explainable artificial intelligence and their applications in  energy systems.
\end{IEEEbiography}

\begin{IEEEbiography}[{\includegraphics[width=1in,height=1.25in,clip,keepaspectratio]{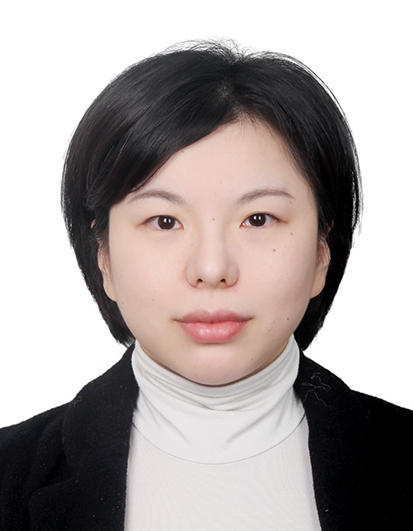}}]{Qiong Huang} received the Ph.D. degree from Okinawa Institute of Science and Technology Graduate University, Japan, in 2023. She is currently a Postdoctoral Researcher with Institute for Automation and Applied Informatics, Karlsruhe Institute of Technology, Germany. Her research interests include reinforcement learning, multi-agent system, and control applications of energy systems.
\end{IEEEbiography}

\begin{IEEEbiography}[{\includegraphics[width=1in,height=1.25in,clip,keepaspectratio]{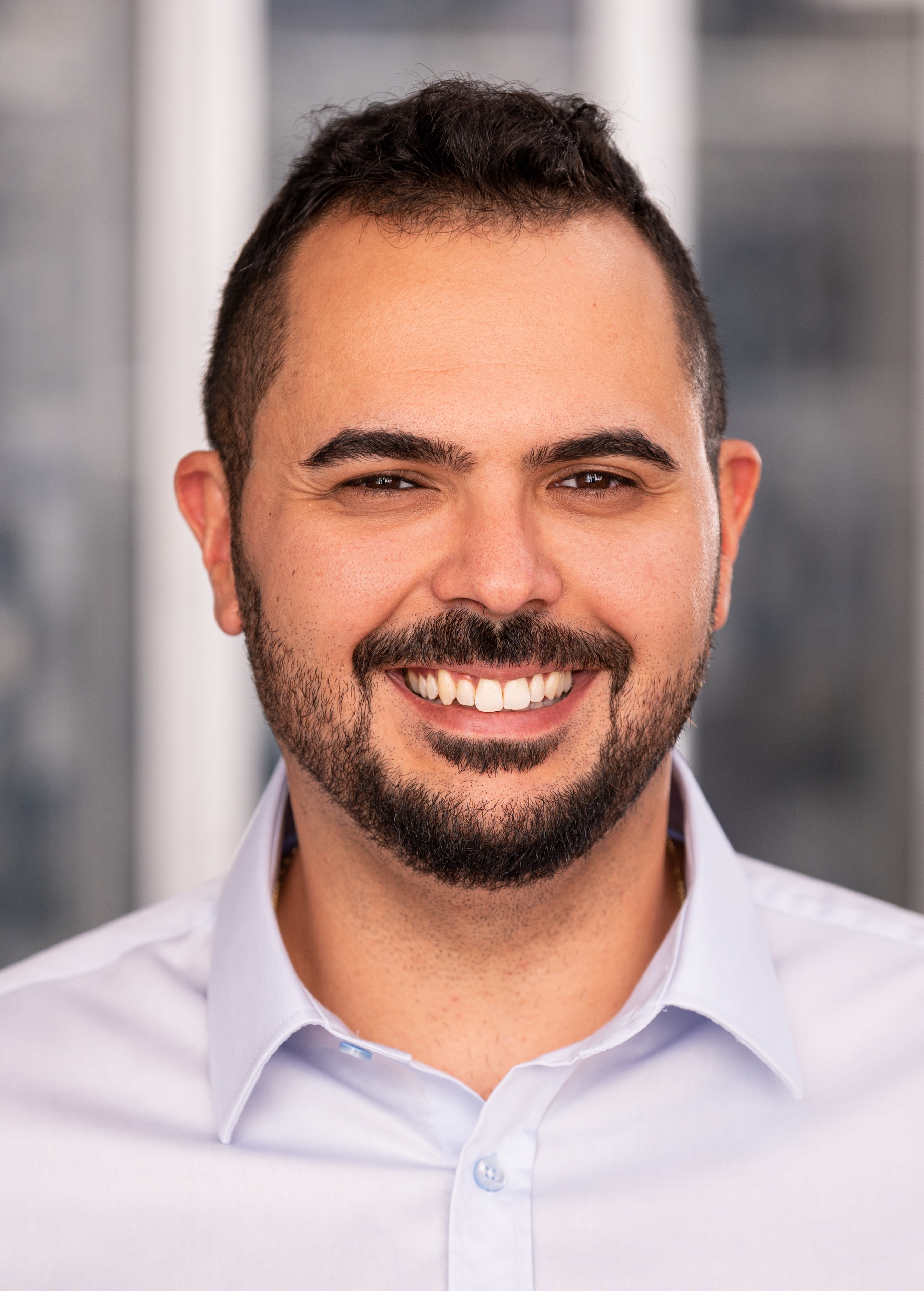}}]{Gökhan Demirel} received the B.Sc. and M.Sc. degrees in electrical engineering and information technology from the Karlsruhe Institute of Technology (KIT), Karlsruhe, Germany, in 2017 and 2021. His research interests include optimal control, smart grid, reinforcement learning, and their applications. 
\end{IEEEbiography}

\begin{IEEEbiography}[{\includegraphics[width=1in,height=1.25in,clip,keepaspectratio]{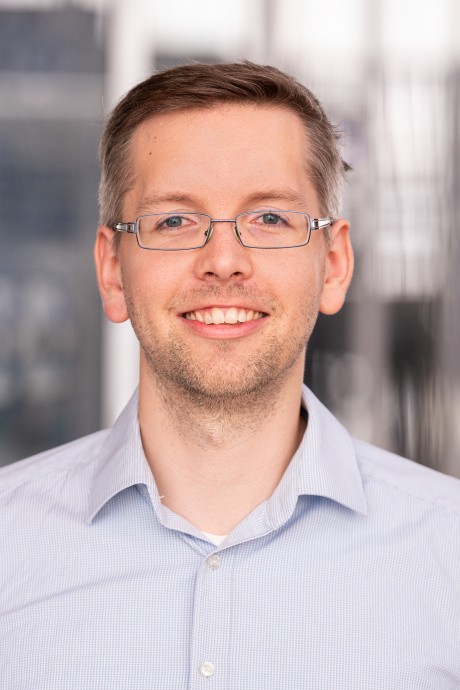}}]{Kevin Förderer} received the B.Sc. degree in industrial engineering and management and the M.Sc. degree in Economathematics from the Karlsruhe Institute of Technology (KIT), Karlsruhe, Germany, before working with FZI Forschungszentrum Informatik as a Research Associate and receiving the Ph.D. degree from KIT in 2021. In 2021, he became the Head of Group for the research group IT Methods and Components for Energy Systems (IT4ES) with the Institute of Automation and Applied Informatics, KIT.
\end{IEEEbiography}

\begin{IEEEbiography}[{\includegraphics[width=1in,height=1.25in,clip,keepaspectratio]{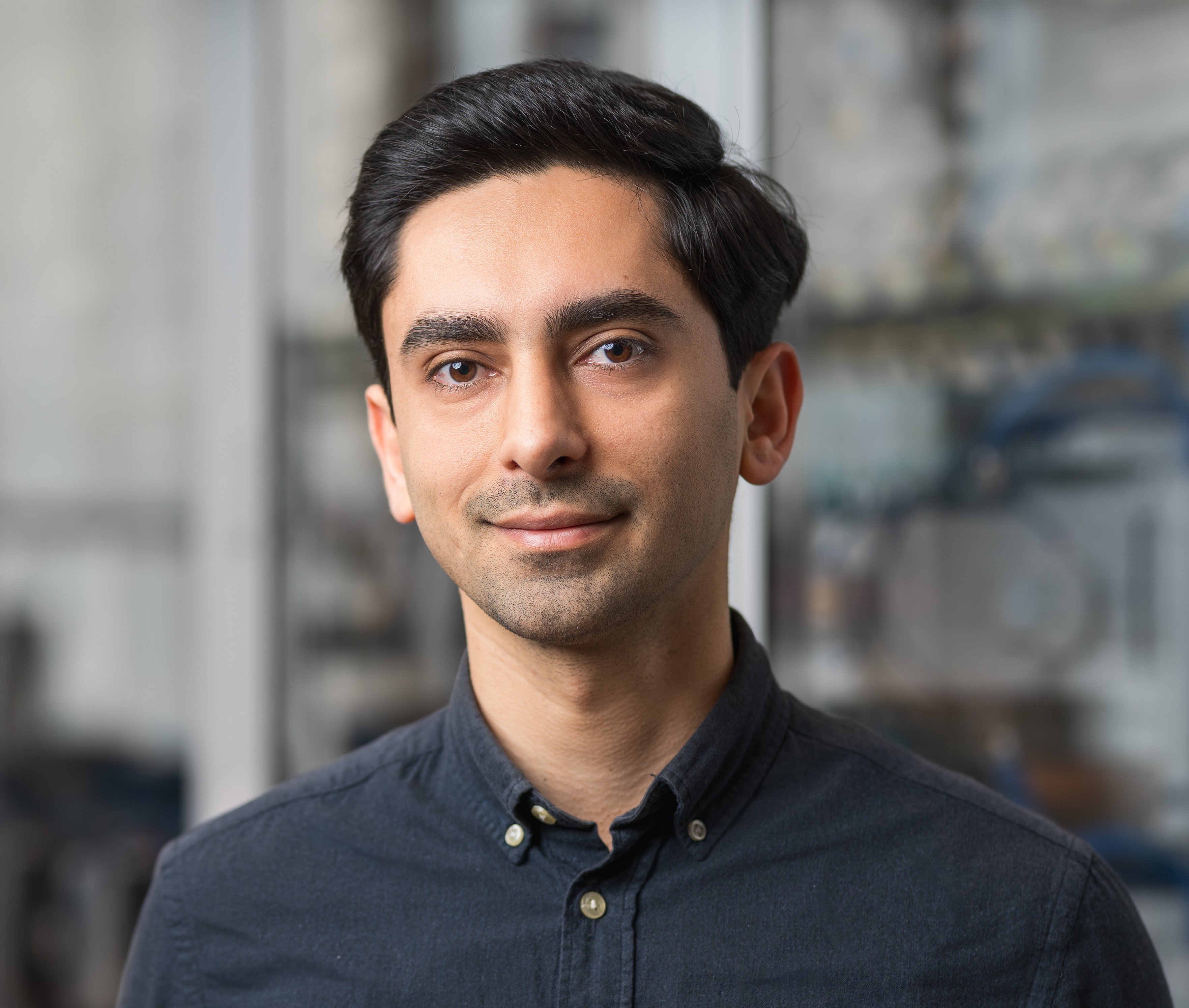}}]{Erfan Tajalli-Ardekani} received the B.Sc. degree in industrial engineering and management and the M.Sc. degree in Economathematics from the Karlsruhe Institute of Technology (KIT), Karlsruhe, Germany, before working with FZI Forschungszentrum Informatik as a Research Associate and receiving the Ph.D. degree from KIT in 2021. In 2021, he became the Head of Group for the research group IT Methods and Components for Energy Systems (IT4ES) with the Institute of Automation and Applied Informatics, KIT.
\end{IEEEbiography}

\begin{IEEEbiography}[{\includegraphics[width=1in,height=1.25in,clip,keepaspectratio]{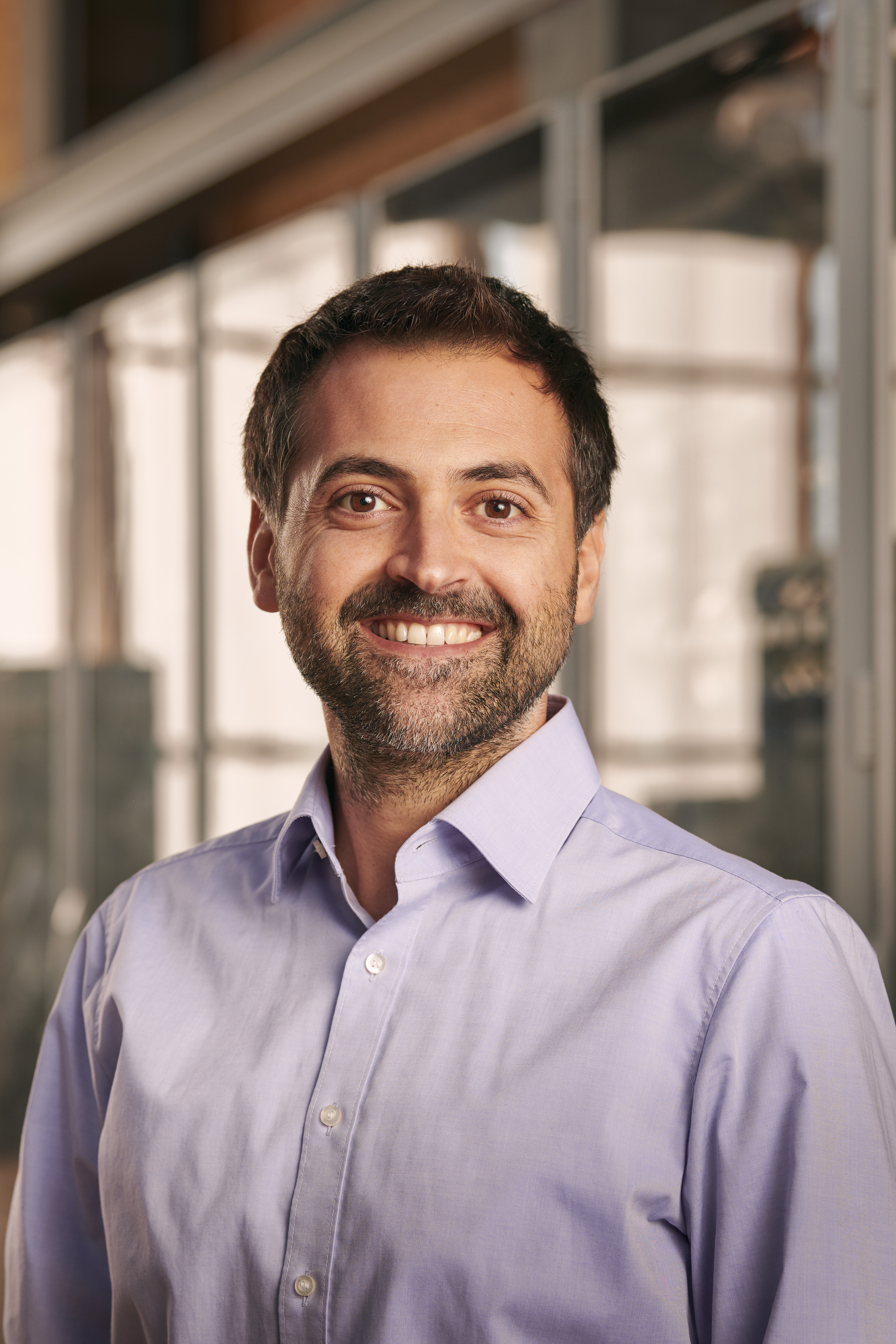}}]{Luigi Spatafora} received his Ph.D. in mechanical engineering from the Karlsruhe Institute of Technology. He leads the Scientific Infrastructure Group at the Energy Lab and the Living Lab Energy Campus. His research interests include renewable energy–based buildings and districts.
\end{IEEEbiography}

\begin{IEEEbiography}[{\includegraphics[width=1in,height=1.25in,clip,keepaspectratio]{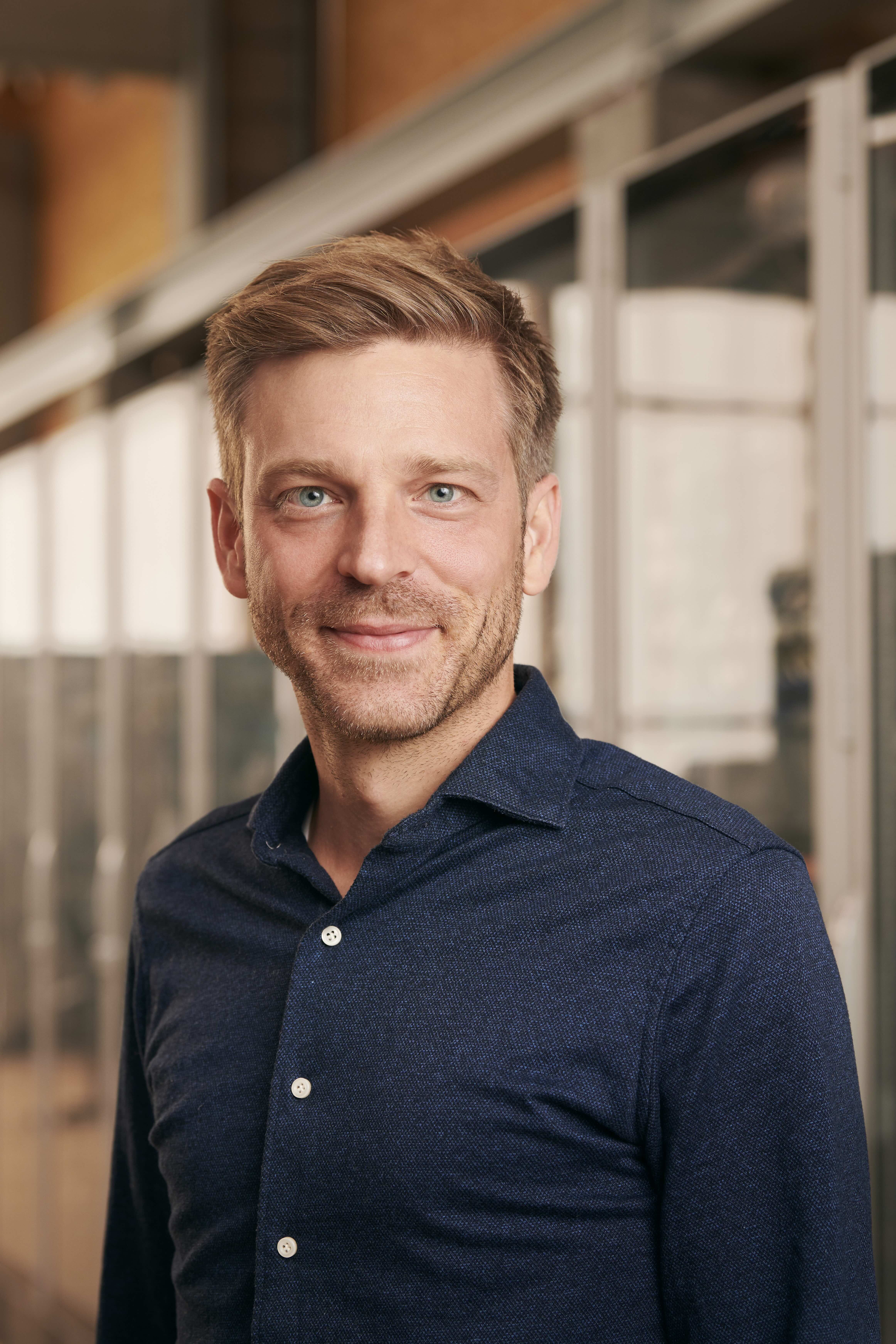}}]{Simon Waczowicz} received the Ph.D. degree in mechanical engineering from KIT in 2018.
Since then, he has headed the Research Platform Energy department at the IAI at KIT. His research
interests include energy system design and operation, and time series analysis and forecasting.

\end{IEEEbiography}

\begin{IEEEbiography}[{\includegraphics[width=1in,height=1.25in,clip,keepaspectratio]{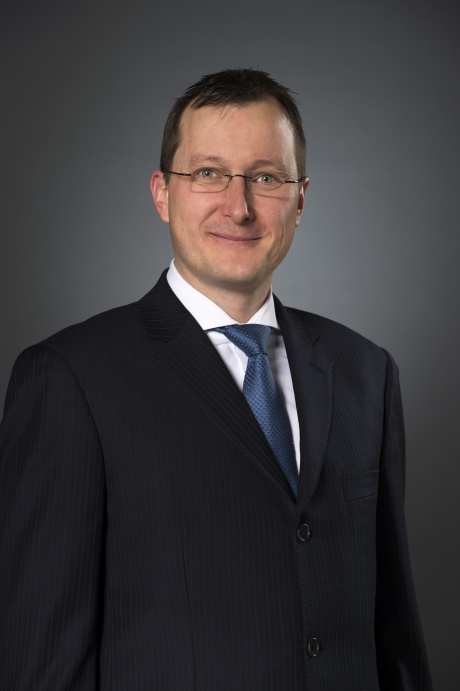}}]{Veit Hagenmeyer} is currently the Professor of Energy Informatics with the Faculty of Informatics, and the Director of the Institute for Automation and Applied Informatics, Karlsruhe Institute of Technology, Karlsruhe, Germany. His research interests include modeling, optimization and control of sector-integrated energy systems, machine learning based forecasting of uncertain demand and production in energy systems mainly driven by renewables, and integrated cyber-security of such systems.

\end{IEEEbiography}

\begin{IEEEbiography}[{\includegraphics[width=1in,height=1.25in,clip,keepaspectratio]{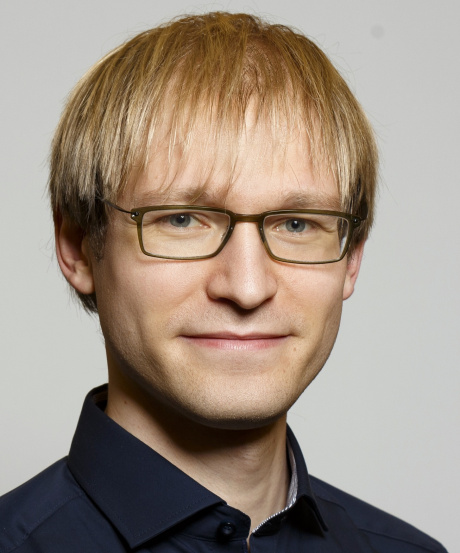}}]{Benjamin Schäfer} is an Assistant Professor at the Department of Informatics at KIT, Germany, since 2023. He investigates power systems with the help of (explainable) machine learning and stochastic modelling. Prior to joining KIT, he worked in Göttingen (Germany), Dresden (Germany, London (United Kingdom), Tokyo (Japan) and Ås (Norway), graduating with a PhD thesis in 2017. 

\end{IEEEbiography}

\EOD

\end{document}